\documentclass[]{bytedance}
\usepackage[toc,page,header]{appendix}
\usepackage{minitoc}
\usepackage{algorithm}
\usepackage{algpseudocode}
\usepackage{booktabs}
\usepackage{multirow}
\usepackage{siunitx}
\usepackage{makecell}
\usepackage{mathtools}
\usepackage{tabularx,array}
\usepackage[table]{xcolor}
\usepackage{graphicx}
\usepackage{wrapfig}
\usepackage{xspace}
\usepackage{amsmath}
\usepackage{amsfonts}
\usepackage{epsfig}
\usepackage{amssymb}
\usepackage{etoolbox}
\usepackage{listings}
\usepackage{enumitem}
\usepackage{amsthm}

\usepackage{graphbox}
\usepackage{hyperref}

\title{\textsc{OmniShow}: Unifying Multimodal Conditions for Human-Object Interaction Video Generation}

\newcommand{\eg}{\textit{e.g.}}
\newcommand{\ie}{\textit{i.e.}}

\newcommand{\method}{\textsc{OmniShow}\xspace}
\newcommand{\bench}{HOIVG-Bench\xspace}
\newcommand{\task}{Human-Object Interaction Video Generation\xspace}
\newcommand{\techVision}{Unified Channel-wise Conditioning\xspace}
\newcommand{\techAudio}{Gated Local-Context Attention\xspace}

\newcommand{\techTrain}{Decoupled-Then-Joint Training\xspace}

\newcommand{\myparagraphF}[1]{\noindent\textbf{#1}}
\newcommand{\myparagraph}[1]{\noindent\textbf{#1}}

\author[1,\star]{Donghao Zhou}
\author[2,\star]{Guisheng Liu}
\author[2]{Hao Yang}
\author[2,\dagger]{Jiatong Li}
\author[3]{\\Jingyu Lin}
\author[4]{Xiaohu Huang}
\author[2]{Yichen Liu}
\author[2]{Xin Gao}
\author[3]{Cunjian Chen}
\author[2,\S]{\\Shilei Wen}
\author[1]{Chi-Wing Fu}
\author[1,\S]{Pheng-Ann Heng}

\affiliation[1]{The Chinese University of Hong Kong}
\affiliation[2]{ByteDance}
\makeatletter
\g@addto@macro\affiliationlist{,\\\affiliationformat[3]{Monash University}, \affiliationformat[4]{The University of Hong Kong}}
\makeatother

\contribution[\star]{Equal contribution}
\contribution[\dagger]{Project lead}
\contribution[\S]{Corresponding author}

\abstract{
In this work, we study \textbf{Human-Object Interaction Video Generation (HOIVG)}, which aims to synthesize high-quality human-object interaction videos conditioned on text, reference images, audio, and pose. 
This task holds significant practical value for automating content creation in real-world applications, such as e-commerce demonstrations, short video production, and interactive entertainment.
However, existing approaches fail to accommodate all these requisite conditions.
We present \textbf{\method}, an end-to-end framework tailored for this practical yet challenging task, capable of harmonizing multimodal conditions and delivering industry-grade performance. 
To overcome the trade-off between controllability and quality, we introduce \textbf{\techVision} for efficient image and pose injection, and \textbf{\techAudio} to ensure precise audio-visual synchronization. 
To effectively address data scarcity, we develop a \textbf{\techTrain} strategy that leverages a multi-stage training process with model merging to efficiently harness heterogeneous sub-task datasets. Furthermore, to fill the evaluation gap in this field, we establish \textbf{\bench}, a dedicated and comprehensive benchmark for HOIVG. 
Extensive experiments demonstrate that \method achieves overall state-of-the-art performance across various multimodal conditioning settings, setting a solid standard for the emerging HOIVG task.
}

\checkdata[Project Page]{\url{https://correr-zhou.github.io/OmniShow}}
\begin{document}
\maketitle

\begin{figure}[t!]
    \centering

    \includegraphics[width=1\linewidth]{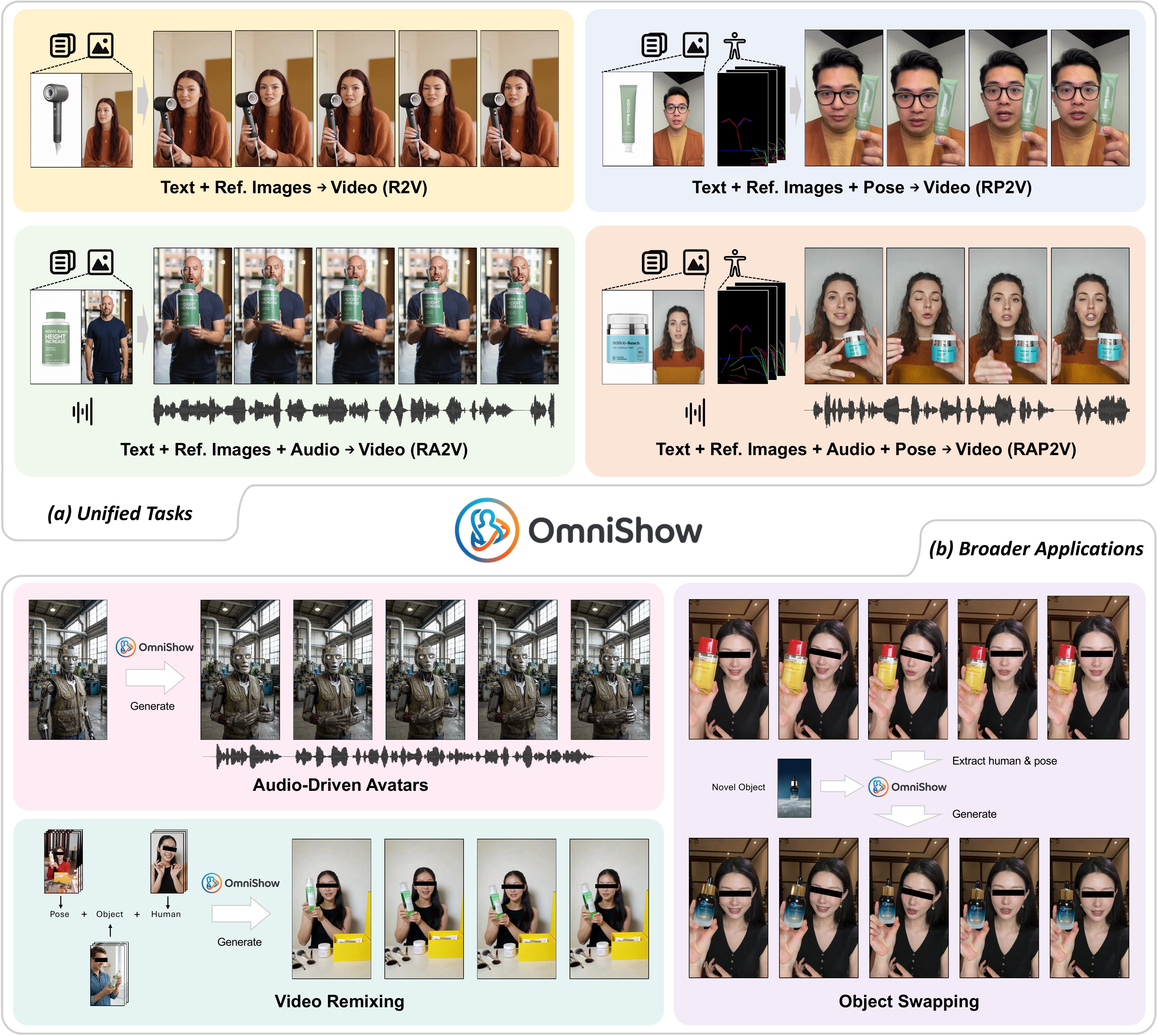}
    
    \captionsetup{hypcap=false}
    \captionof{figure}
    { 
        \textbf{Overview of \method.}
        Our model unifies text, reference image, audio, and pose conditions to synthesize high-quality videos.
        \textit{(a) Unified Tasks:} \method exhibits exceptional flexibility across diverse conditioning settings, from R2V, RA2V, RP2V to the most challenging RAP2V, consistently delivering industry-grade visual quality and precise multimodal alignment.
        \textit{(b) Broader Applications:} \method unlocks creative potential such as audio-driven avatars, object swapping, and video remixing, demonstrating robust compositional control and extensible capabilities.
        \textit{Please visit our project page for more immersive and diverse video demonstrations.}
    }
    
    \label{fig:teaser}
    
\end{figure}

\section{Introduction}
\label{sec:intro}

The rapid evolution of text-to-video generation models \cite{wan2025wan, zhang2025waver, wu2025hunyuanvideo, chen2025seedance} has revolutionized content creation, especially in human-centric scenarios. 
While human-centric video generation has achieved impressive visual fidelity, real-world applications—such as e-commerce demonstrations, short video production, and interactive entertainment—demand precise controllability over specific subjects and their dynamics. 
We term this practical yet challenging task as \textbf{Human-Object Interaction Video Generation (HOIVG)}. Specifically, it requires synthesizing high-quality videos conditioned on four distinct inputs: a \textit{text} prompt for global semantics, \textit{reference images} for specific character and object appearance, \textit{audio} for synchronized lip and body movements, and a \textit{pose} sequence for explicit motion control.

Despite the progress, existing methods struggle to unify the diverse multimodal conditions required for HOIVG.
Reference-to-Video (R2V) approaches \cite{liu2025phantom,fei2025skyreels} focus on subject preservation but lack audio responsiveness, resulting in ``silent'' interactions.
Conversely, Audio-to-Video (A2V) methods \cite{cui2025hallo3,gan2025omniavatar,kong2025let} aim for audio synchronization but only support an initial frame rather than reference images, limiting their applicability.
While existing works \cite{hu2025hunyuancustom, chen2025humo} attempt to combine audio and reference images, they overlook pose conditions, which are critical for realizing complex interactions that text cannot fully specify. Moreover, these methods are not tailored for HOIVG, often falling short of high-fidelity subject preservation required by this task.
Although some approaches target this task \cite{xu2024anchorcrafter,wang2025dreamactor,hu2025hunyuancustom}, they rely on mandatory inputs such as predefined object masks or trajectories and lack audio-driven abilities.
Furthermore, recent joint audio-video generation models \cite{huang2025jova, low2025ovi, hacohen2026ltx} primarily rely on text prompts, offering limited customization capabilities.
Consequently, there is currently no unified framework capable of harmonizing text, reference image, audio, and pose conditions in an end-to-end manner, all of which are required by the HOIVG task.

Achieving this unification presents three primary challenges:
(i) A critical trade-off exists between multimodal controllability and generation quality. 
Naively introducing aggressive modifications to handle multimodal inputs typically disrupts the base model's pretrained generative priors.
(ii) High-quality HOI datasets containing paired quintuplet data (conditions and the target video) are scarce. Instead, readily available resources are fragmented into isolated sub-tasks (\eg, separate A2V datasets).
(iii) The community currently lacks a dedicated and comprehensive benchmark to evaluate HOIVG under such diverse multimodal conditions, hindering sustainable research exploration in this field.

In this work, we propose \textbf{\method}, an end-to-end framework designed for HOIVG (\cref{fig:pipeline}). 
By simultaneously orchestrating text, reference image, audio, and pose conditions, \method unlocks the full potential of multimodal control while achieving superior generation quality (\cref{fig:teaser}).
Our methodology is driven by a triad of efforts: \textit{integrating efficient and pragmatic techniques, synergistically utilizing heterogeneous training data, and establishing a standardized benchmark}. 

First, we introduce \textbf{\techVision} (\cref{fig:pipeline}a).
We augment the noisy video tokens by padding additional tokens (termed “pseudo-frame tokens”) along the temporal dimension.
Both the pose video and reference images are encoded by VAE, and then injected via the same channel-concatenation strategy:
the pose tokens are concatenated with the noisy video tokens, and the image tokens with the pseudo-frame tokens.
Furthermore, we employ a reconstruction loss on pseudo-frames to encourage the preservation of reference images' semantic details.
This mechanism maintains the native input structure and token distribution of the base model, thereby minimizing the adaptation gap and achieving efficient, seamless injection.

Second, we design \textbf{\techAudio} (\cref{fig:pipeline}b).
An audio context packing strategy is leveraged to aggregate rich audio features that encapsulate sufficient contextual information.
To achieve temporal alignment, we employ a masked attention mechanism that restricts video tokens to interact only with their corresponding audio segments.
Furthermore, we incorporate learnable gating vectors to modulate the injection, which stabilizes early training and serves as an explicit indicator for the impact of audio cues.
Collectively, these design choices facilitate precise audio-visual synchronization in generated videos.

Third, we develop \textbf{\techTrain} (\cref{fig:pipeline}c).
We start by establishing multiple data pipelines to collect, filter, and organize diverse training data for HOIVG. 
To fully leverage these heterogeneous data, we initially train separate A2V and R2V models, allowing each to specialize in its respective modality. 
These models are then fused for joint training to harmonize text, reference images, and audio.
Next, pose is introduced in the final stage to prevent over-reliance on this strong signal.
Such a strategy not only boosts data utility but also validates that fusing specialized models benefits robust multimodal training.

For a systematic evaluation, we propose \textbf{\bench}, a dedicated and comprehensive benchmark for HOIVG. 
Experiments on this benchmark show that \method delivers superior or competitive performance against existing R2V, A2V, and other methods in multiple multimodal control settings. 
We believe that \method will set a solid standard for the emerging HOIVG task, and the methodological and empirical insights presented herein could inspire future advancements in video generation.

Our main contributions are summarized as follows:
\begin{itemize}[itemsep=-0.2em, topsep=-0.2em]
    \item We propose \method, the first-of-its-kind framework for HOIVG, capable of harmonizing multimodal conditions. To achieve this, we introduce \textit{\techVision} and \textit{\techAudio}, which enable precise controllability without compromising generation quality.
    \item We develop a \textit{\techTrain} strategy, which leverages a multi-stage training process with model merging to efficiently harness heterogeneous data from diverse sub-task datasets, effectively circumventing the data scarcity of the HOIVG training.
    \item We establish \textit{\bench}, a dedicated and comprehensive benchmark for HOIVG evaluation. Extensive experiments on \bench validate that \method achieves state-of-the-art performance in various multimodal conditioning settings.
\end{itemize}

\section{Related Work}
\label{sec:rw}

\myparagraphF{Controllable Video Generation} aims to synthesize videos conditioned on diverse inputs beyond text.
Recent advances in visual generative models have motivated extensive research, from conditional image synthesis \cite{zhou2026identitystory, lin2025jarvisevo, liu2026hifi, zhou2024magictailor, song2025scenedecorator, qin2026innoads, chen2026posteromni, huang2025dual} to controllable video generation \cite{huang2025m4v, huang2024magicfight, song2025hero, wang2025language, ling2026mofu, wang2025wisa, shao2025magicdistillation}.
Reference-to-Video (R2V) generation, also referred to as video customization, focuses on preserving the subject identity of input reference images \cite{yuan2025identity,liu2025phantom, fei2025skyreels, zhou2025scaling}.
Driven by the demand for digital avatars, Audio-to-Video (A2V) generation has evolved from talking heads \cite{zhang2023sadtalker, jiang2024loopy} to portrait animation \cite{cui2025hallo4, lin2025omnihuman, linapoavatar, yi2025magicinfinite, weimocha} and multi-person conversation \cite{zhong2025anytalker, wang2025interacthuman, kong2025let}.
Concurrently, pose-guided approaches leverage explicit structural signals, ranging from skeleton maps \cite{xue2024follow, gan2025humandit} to dense correspondences \cite{xu2024magicanimate}, to direct the generation of human motion.
Recently, there is a growing trend towards integrating multiple conditions \cite{hu2025hunyuancustom, jiang2025vace, chen2025humo, team2025kling}.
However, establishing a robust framework that collaborates text, reference image, audio, and pose conditions, which are required by our studied task, remains a significant open challenge.

\myparagraph{\task (HOIVG)} focuses on synthesizing realistic and vivid HOI videos grounded in multimodal conditions. 
Prior research on HOI has spanned in 3D reconstruction \cite{xu2021d3d, ye2023diffusion} and motion sequence synthesis \cite{diller2024cg, peng2025hoi}.
With the advancement of visual generative models, previous works have begun to investigate HOI in image generation \cite{xue2024hoi, chen2024virtualmodel, Fan_2025_CVPR}.
Recently, HOIVG has emerged as a prominent topic, driven by its critical role in real-world applications.
AnchorCrafter \cite{xu2024anchorcrafter} utilizes body skeletons, hand meshes, and object depth maps to guide the interaction;
HunyuanVideo-HOMA \cite{huang2025hunyuanvideo} proposes to use sparse human poses and object trajectory dots; 
and DreamActor-H1 \cite{wang2025dreamactor} rely on body mesh templates and object bounding boxes.
However, these methods are all constrained by strict input requirements and cannot utilize audio cues, struggling to achieve satisfactory flexibility and generation quality.
In contrast, our \method supports flexible configurations of multimodal conditions, while delivering outstanding performance.
\section{Methodology}
\label{sec:method}

\begin{figure*}[t]
    \centering
    
    \includegraphics[width=1\linewidth]{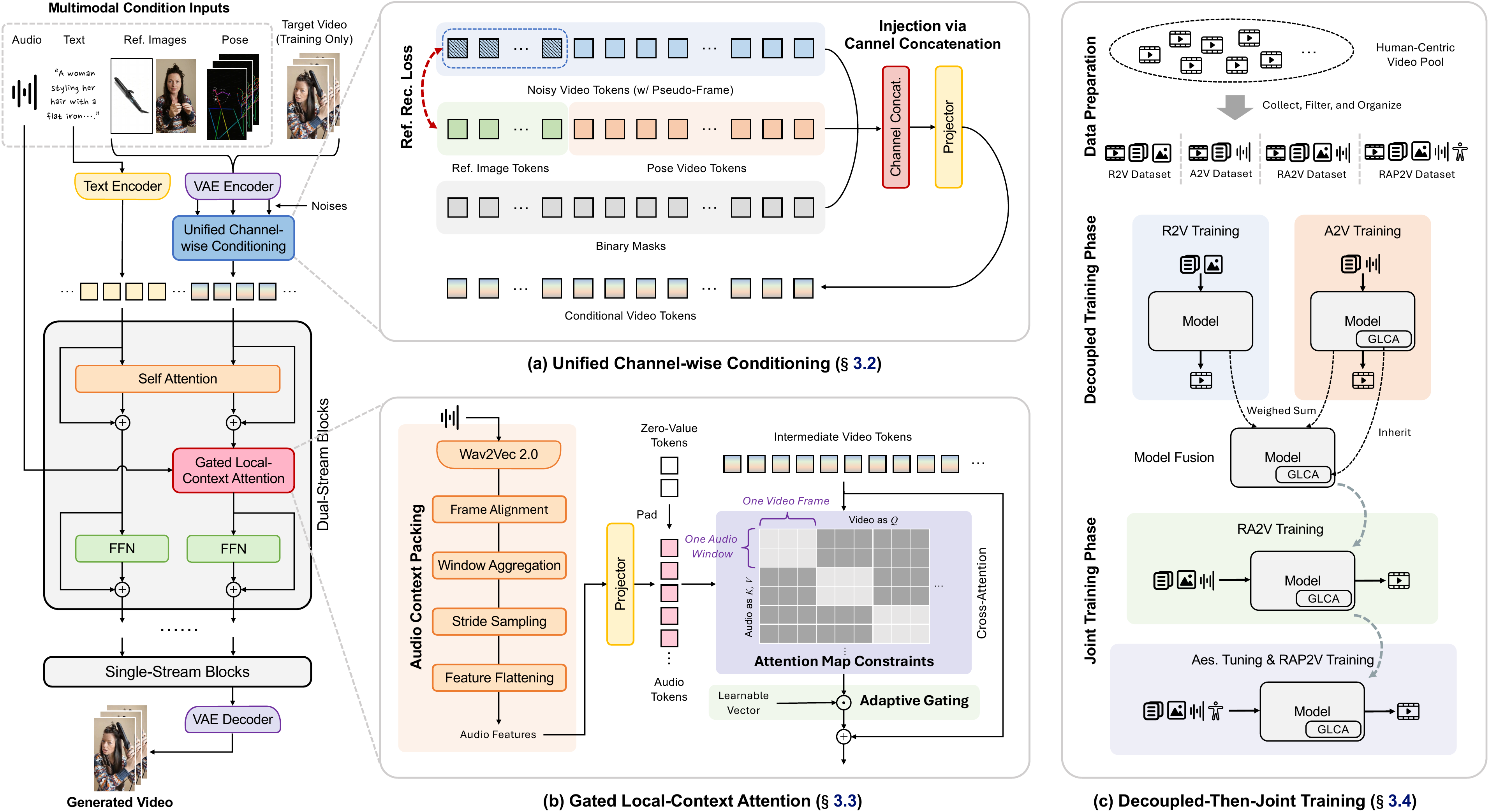}

    \caption
    { 
        \textbf{Pipeline of \method.}
Our framework consists of:
\textit{(a) \techVision} (\cref{sec:tech_vison}) effectively injects reference images and pose cues via unified channel concatenation. It augments noisy video tokens with pseudo-frames, which are supervised by a reference reconstruction loss to preserve semantic details.
\textit{(b) \techAudio}  (\cref{sec:tech_audio}) ensures precise audio-visual synchronization. It packs raw audio features with sufficient contextual information and injects them via masked attention to align video frames with corresponding audio segments, followed by adaptive gating to stabilize early training.
\textit{(c) \techTrain}  (\cref{sec:tech_train}) makes the efficient utilization of heterogeneous datasets possible. We first train specialized R2V and A2V models on separate sub-task datasets, then fuse them via weight interpolation, followed by joint fine-tuning to unify multimodal capabilities.
Note that we omit some components, such as normalization, within the MMDiT blocks for brevity.
    }

    \label{fig:pipeline}
    
\end{figure*}

As illustrated in \cref{fig:pipeline}, our \method generates high-quality videos by conditioning on a flexible combination of multimodal inputs.
Built upon Waver 1.0 \cite{zhang2025waver} (\cref{sec:prel}), a powerful 12B MMDiT-based model, the framework comprises four key components: 
\textit{\techVision} (\cref{sec:tech_vison}) for efficiently injecting reference images and pose cues without disrupting the generative priors; 
\textit{\techAudio} (\cref{sec:tech_audio}) for ensuring precise synchronization between audios and human dynamics; 
\textit{\techTrain} (\cref{sec:tech_train}) for effectively harnessing heterogeneous datasets;
and \textit{\bench} (\cref{sec:bench_details}) for providing a comprehensive evaluation suite to systematically assess this task.

\subsection{Preliminary}
\label{sec:prel}
Waver 1.0 follows the latent diffusion paradigm, utilizing Wan 2.1 VAE \cite{wan2025wan} to compress video features as latent tokens.
In the latent space, Flow Matching \cite{lipman2022flow} is adopted for training, where the objective is to minimize the discrepancy between the predicted velocity field $v_\theta$ and the ground-truth flow velocity $u$:
\begin{equation}
    \mathcal{L}_{\text{FM}} = 
    \mathbb{E}_{t, \mathbf{x}_0, (\mathbf{x}_1, \mathbf{e})} \left[ \| v_\theta(t, \mathbf{x}_{\text{in}}, \mathbf{e}) - u(\mathbf{x}_t | \mathbf{x}_1) \|^2 \right], 
\end{equation}
where $t$ represents the timestep, $\mathbf{x}_0 \sim \mathcal{N}(0, \mathbf{I})$ denotes the Gaussian noise, and $\mathbf{x}_1$ is the clean video tokens with $\mathbf{e}$ as the paired text embedding. 
The noisy video tokens $\mathbf{x}_t = \mathbb{R}^{N \times D}$ contains $N$ tokens with channel dimension $D$, and $u(\mathbf{x}_t | \mathbf{x}_1) = \mathbf{x}_1 - \mathbf{x}_0$ is the corresponding flow velocity.

Moreover, Waver 1.0 is a task-unified model supporting both the Text-to-Video (T2V) and Image-to-Video (I2V) tasks. To achieve this, it expands $\mathbf{x}_t$ via channel concatenation to form the model input:
\begin{equation}
\mathbf{x}_{\text{in}} = \text{Concat}(\mathbf{x}_t, \mathbf{c}, \mathbf{m}),
\end{equation}
where $\mathbf{c} \in \mathbb{R}^{N \times D}$ represent the condition tokens, and $\mathbf{m} \in {[0, 1]}^{N \times 4}$ is the binary mask indicating the condition status.
Specifically, for the T2V task, $\mathbf{c}$ is filled by black image tokens, and $\mathbf{m}$ is set to all zeros. 
As for the I2V task, to condition on the first frame, the corresponding part of $\mathbf{c}$ is replaced by the input first-frame image tokens, and the corresponding entries in $\mathbf{m}$ are set to 1.

\begin{figure}[t]
    \centering
    \begin{minipage}[t]{0.45\linewidth}
        \centering
        
        \includegraphics[width=1\linewidth]{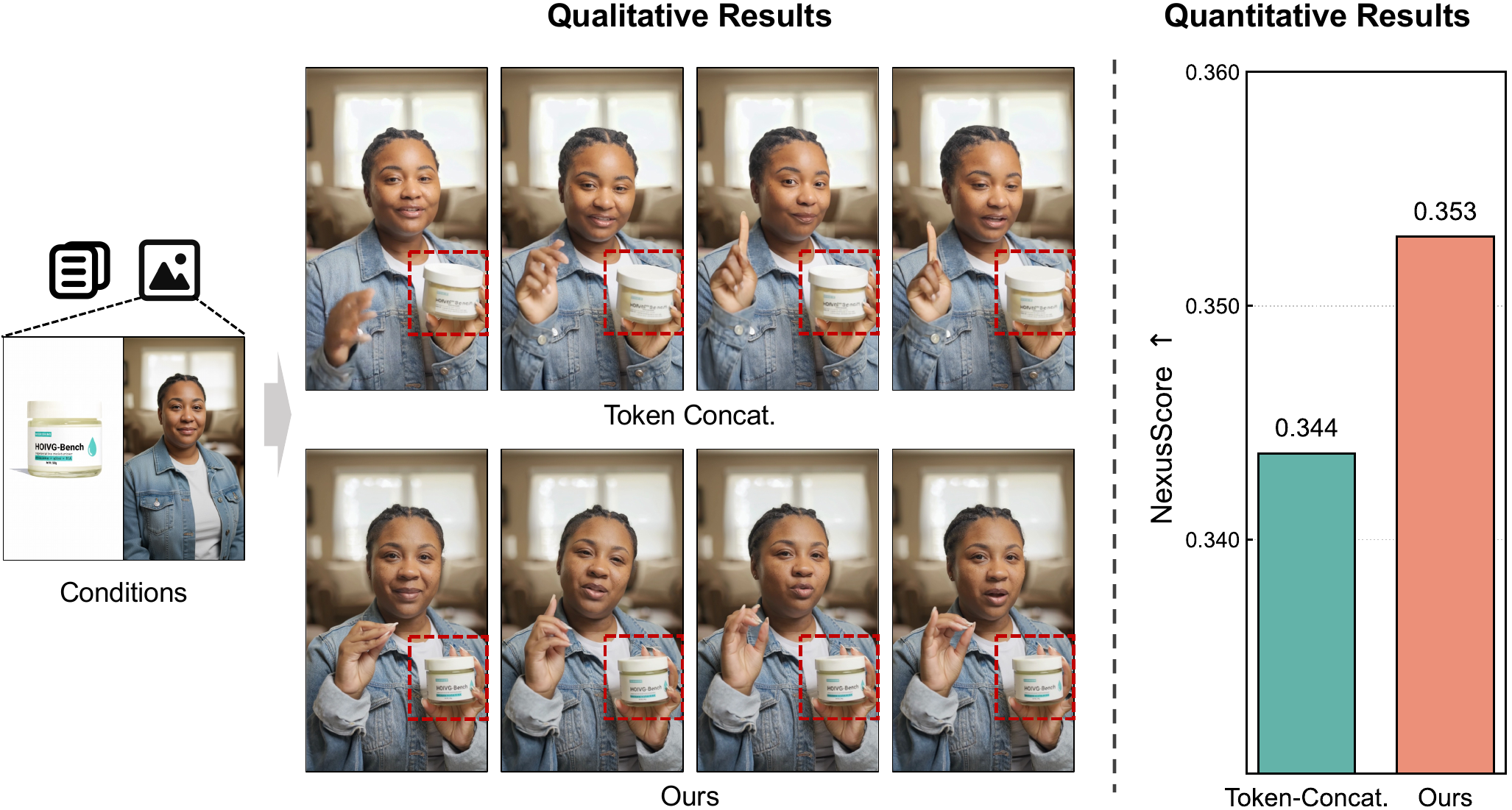}
        
        \caption
        { 
            \textbf{Comparison of conditioning methods}, including a qualitative example and quantitative results based on the HOIVG-Bench.
            The proposed mechanism accurately preserves the visual appearance of the reference subject.
        }
        
        \label{fig:channel_concat}
    \end{minipage}
    \hfill
    \begin{minipage}[t]{0.53\linewidth}
        \centering
        
        \includegraphics[width=1\linewidth]{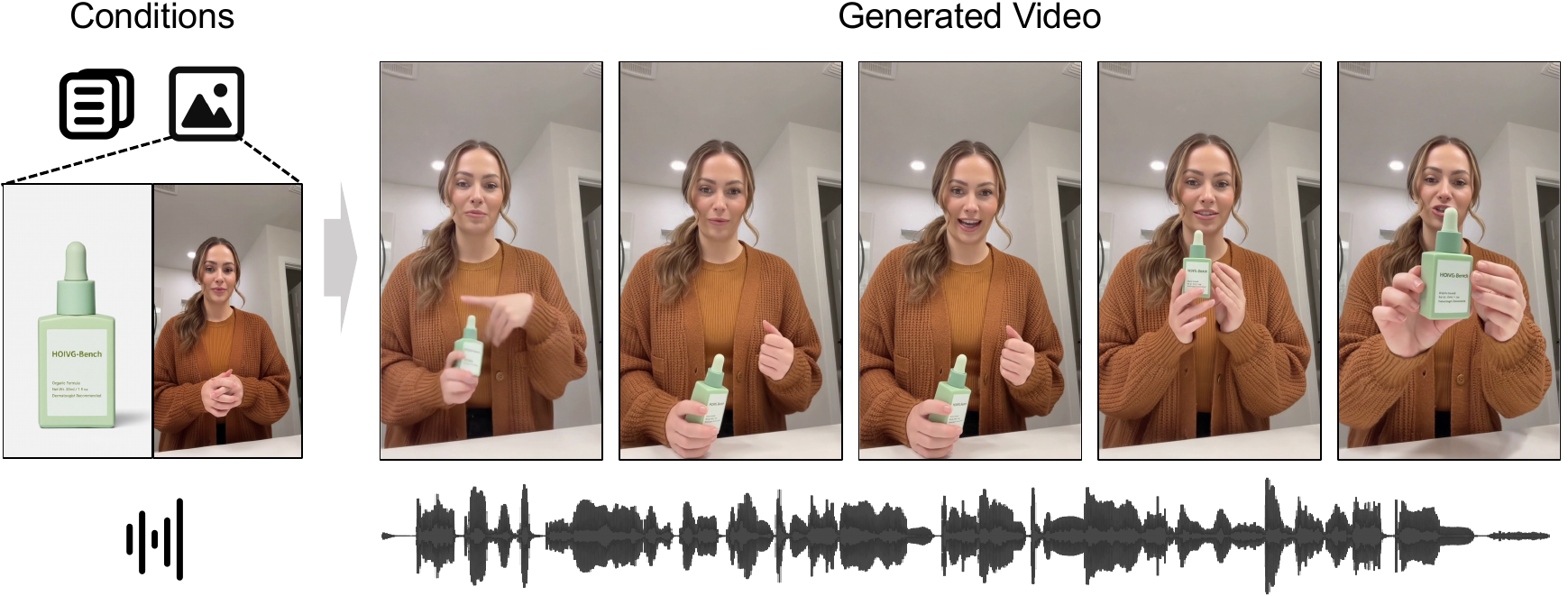}
        
        \caption
        { 
        \textbf{Zero-shot RA2V generation via model merging.}
        Despite not being explicitly trained for the RA2V task yet, the merged model successfully generates videos that respect both reference images and audio inputs, which demonstrates the effectiveness of our strategy.
        }
        \vspace{-3mm}
        
        \label{fig:ra2v_init_results}
    \end{minipage}
\end{figure}

\subsection{\techVision}
\label{sec:tech_vison}

\myparagraph{Injection via Channel Concatenation.}
To harmonize the injection of reference images and pose while maximally preserving the generation quality, we follow and extend the native conditioning paradigm.
First, the pose sequence is rendered as an RGB video, and pose video tokens $\mathbf{p} \in \mathbb{R}^{N \times D}$, reference image tokens $\mathbf{r} \in \mathbb{R}^{N' \times D}$ are obtained by VAE encoding. 
Originally, $\mathbf{x}_t$ only provides $N$ conditioning slots. 
To structurally accommodate the joint injection, we expand this capacity by augmenting $\mathbf{x}_t$ with $N'$ pseudo-frame tokens $\mathbf{x}' \in \mathbb{R}^{N' \times D}$. 
Consequently, the conditions are injected via a unified channel-wise concatenation strategy:
\begin{equation}
\mathbf{x}_{\text{in}} = \text{Concat}([\mathbf{x}' \parallel \mathbf{x}_t], [\mathbf{r}  \parallel \mathbf{p}], [\mathbf{m}' \parallel \mathbf{m}]),
\end{equation}
where $\parallel$ denotes concatenation along the temporal axis, and $\mathbf{m}' \in \mathbb{R}^{N' \times D}$ is the corresponding augmented mask. 
This unified design allows the model to efficiently assimilate the global appearance reference with temporally aligned pose cues simultaneously.

\myparagraph{Reference Reconstruction Loss.}
Leaving pseudo-frame tokens $\mathbf{x}'$ as all-zero tensors provides no informative guidance for the model.
To improve this, we initialize $\mathbf{x}'$ with the noisy reference image tokens perturbed by the same timestep $t$, and enforce a Flow Matching loss $\mathcal{L}_{\text{FM-ref}}$ to facilitate the reconstruction of reference images, with the loss weight set to 1.
This strategy ensures synchronized denoising dynamics and explicitly compels the model to perceive and retain high-fidelity semantic details, thereby ensuring robust visual consistency with the conditional reference images.

\myparagraph{Advantage Analysis.}
While the token concatenation strategy has proven effective in similar tasks \cite{tan2025ominicontrol, ju2025fulldit}, our method offers superior performance for task-unified video generation models (\cref{fig:channel_concat}\&\cref{tab:abl_tech_vision}).
We attribute this to \textit{the minimization of the task adaptation gap}: 
instead of introducing hybrid tokens that bring substantial learning costs, we preserve the native input structure for conditioning, allowing the model to transfer the pretrained I2V capability efficiently.
Such a design highlights the potential of advanced channel-wise conditioning as a worth-exploring solution to controllable video generation.

\subsection{\techAudio}

\label{sec:tech_audio}

\myparagraph{Audio Context Packing.}
We adopt an effective scheme to consolidate rich audio features from raw audio signals. 
First, the audio is fed into Wav2Vec 2.0 \cite{baevski2020wav2vec}, where representations from multiple layers are merged to capture both semantic and rhythmic attributes. 
Then, linear interpolation is employed to match the fps of the original video. 
To aggregate temporal context, we adopt a sliding window strategy with a size of $w = 5$, stacking neighbors for each audio feature along an extra dimension. 
These features are then sampled with a stride of $s = 4$ to align with the VAE temporal compression. 
Finally, the contextual features are flattened in chronological order, yielding dense 2D features rich in contextual information.

\myparagraph{Attention Map Constraints.}
The packed features are processed by a shared audio projector and then interact with video tokens through cross-attention. 
To achieve precise audio-visual temporal alignment, we constrain each latent frame's video tokens to attend only with its corresponding $w$ audio tokens, resulting in a masked attention mechanism:
\begin{equation}
\text{Attn}(Q, K, V, M) = \text{softmax}\left(\frac{QK^T}{\sqrt{d_k}} + \log M\right) V,
\label{eq:attn}
\end{equation}
where queries $Q$ are derived from video tokens, while keys $K$ and values $V$ come from audio tokens. 
$M$ is a binary matrix, where an entry of 1 allows interaction and 0 prevents it (here $\log 0$ is implemented as a large negative constant in practice). 
Notably, we zero-pad audio tokens to align with additional pseudo-frame tokens, a trick essential for accommodating reference image injection.
By eliminating interference from irrelevant audio segments, this attention design enforces fine-grained correspondence between modalities, significantly enhancing audio-visual synchronization.

\begin{wrapfigure}{r}{0.4\columnwidth}
    \vspace{-6mm}
    
    \centering
    
    \includegraphics[width=1\linewidth]{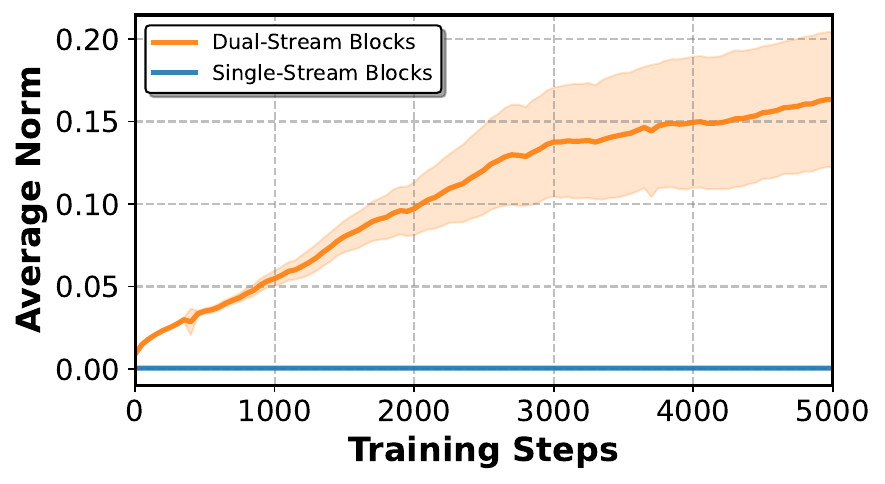}
    
    \vspace{-2mm}
    
    \caption
    { 
    \textbf{Variation of the average norm of $\mathbf{g}$},
    which reflects audio impact in different blocks.
    }
    
    \label{fig:gating_insight}
    
    \vspace{-5mm}
    
\end{wrapfigure}
\myparagraph{Adaptive Gating.}
Directly inserting newly initialized modules can disrupt pretrained feature distributions during early training \cite{zhang2023adding}. 
To avoid this, we introduce a learnable gating vector $\mathbf{g} \in \mathbb{R}^H$, initialized to a near-zero value of $1e-5$, where $H$ is the token hidden dimension. 
With the modulation of $\mathbf{g}$, audio injection is formulated as:
\begin{equation}
\mathbf{h}_o = \mathbf{h}_i + \mathcal{F}_{\text{Attn}}(\mathbf{h}_i, \mathbf{a}) \odot \mathbf{g} ,
\end{equation}
where $\mathbf{h}_i, \mathbf{h}_o \in \mathbb{R}^{(N' + N) \times H}$ denote input and output video tokens, $\mathbf{a}\in \mathbb{R}^{N_a \times H}$ is audio tokens with token number $N_a$,
$\mathcal{F}_{\text{Attn}}$ represents the complete attention operation including \cref{eq:attn} and associated projections, and $\odot$ is element-wise multiplication. 
We argue that this method offers distinct advantages over simply zero-initializing weights within $\mathcal{F}$.
Beyond ensuring training stability, $\mathbf{g}$ also acts as an explicit indicator to guide our architectural design via revealing the magnitude of audio impact. 
Specifically, motivated by the empirical observation of $\mathbf{g}$ (\cref{fig:gating_insight}), we insert audio attention only into the dual-stream blocks for efficient injection. 
This strategic placement merely increases the model scale by \textit{$\sim$2.5\% (totaling 12.3B)}, while existing methods like HuMo \cite{chen2025humo} increase model parameters by \textit{$\sim$21.4\% (totaling 17B)}.

\subsection{\techTrain}
\label{sec:tech_train}

\myparagraph{Training Data.}
The effectiveness of video generation critically relies on training data. 
However, high-quality HOIVG data is scarce, as a video requires valid text, reference images, audio, and pose conditions; failure in any condition leads to discarding the sample. 
Thus, we propose to collect heterogeneous data that meets specific sub-task standards for training. 
We first collect a massive human-centric video pool. 
Then, we build multiple pipelines to construct R2V, A2V, and Reference+Audio-to-Video (RA2V) datasets. 
Consequently, we curate a high-quality Reference+Audio+Pose-to-Video (RAP2V) subset for final fine-tuning.
More details of data collection are included in \cref{sec:data_details}.

\myparagraph{Decoupled Training Phase.}
We start by training specialized R2V and A2V models, utilizing their dedicated datasets (\cref{fig:pipeline}c). 
Note that, for R2V training, we discard audio modules to retain the same architecture as the base model; for A2V training, we follow the common paradigm to incorporate the first-frame image as an additional condition.
Such a phase allows for the efficient utilization of heterogeneous sub-task data. 
Furthermore, it enables each model to master its specific modality conditioning, ensuring robust alignment before subsequent joint training.

\myparagraph{Joint Training Phase.}
Subsequently, we merge the models by inheriting audio modules from the A2V model and linearly interpolating the rest (\cref{fig:pipeline}c), with $0.6, 0.4$ for the A2V and R2V models, respectively.
The ratio selection is guided by a principled observation: audio synchronization (relying on fine-grained temporal alignment) is significantly more sensitive to weight disruption than visual identity (relying on global appearance features).
Remarkably, \textit{the merged model exhibits emergent RA2V capabilities} (\cref{fig:ra2v_init_results}), evidencing the effectiveness of our strategy. 
It is then trained on the full RA2V dataset, followed by further fine-tuning on a high-quality subset to enhance naturalness and aesthetics. 
Notably, as a strong supervision signal, pose is introduced only in the final fine-tuning stage to prevent overfitting. 
Through this phase, we ultimately achieve a unified model capable of multimodal conditioning.

\subsection{\bench}
\label{sec:bench_details}
To systematically evaluate the capabilities of HOIVG under diverse multimodal conditions, we present \textbf{HOIVG-Bench}. 
Existing benchmarks often focus on limited-modality control (\eg, text+pose or text+images), lacking a comprehensive assessment of the synergy between text, images (both human and object), audio, and pose signals, which are required by the HOIVG setting. HOIVG-Bench aims to bridge this gap by providing an evaluation suite comprising 135 carefully curated samples and dedicated metrics. 

\myparagraph{Sample Construction.}
In \bench, each sample is equipped with a detailed textual caption, a human reference image, an object reference image, semantically aligned audio, and a coherent pose sequence.
We adopt a rigorous data curation and processing pipeline to construct the benchmark:

\begin{itemize}[itemsep=-0.2em, topsep=-0.2em]
    \item \textbf{Video Curation:} We initially select raw video from an in-house video library. The selection criteria include: (1) video duration exceeding 4 seconds; (2) the presence of clear human-object interactions; and (3) a diverse range of human attributes (\eg, gender, age, ethnicity) and object categories (\eg, daily necessities, tools) to ensure data diversity.
    
    \item \textbf{Object Image Acquisition:} To simulate real-world generation scenarios, we avoid simply cropping objects from the video. Instead, we utilize Nano Banana \cite{google_nanobanana} to modify the original objects' textures and colors while adding sufficient fine-grained details, resulting in high-quality reference object images.
    
    \item \textbf{Human Image Acquisition:} Considering privacy protection and identity de-identification, we generate reference human images based on video screenshots using Nano Banana. These generated images maintain stylistic similarity to the original subjects while altering identity features, ensuring compliance and testing generalization.

    \item \textbf{Pose Extraction:} We employ DWPose \cite{yang2023effective} to extract per-frame human pose skeletons from the original videos, serving as the ground truth signal for motion control.
    
    \item \textbf{Audio Synthesis:} To construct semantically consistent audio inputs, we design a two-stage generation process. First, GPT-4o \cite{hurst2024gpt} is utilized to generate a speech script focused on describing the target object. Subsequently, GPT-4o analyzes the gender and age attributes of the human reference image, and ElevenLabs \cite{elevenlabs_ai_voice} is invoked to synthesize high-quality speech audio with matching timbres.
\end{itemize}

\begin{figure*}[t]
    \centering
    
    \includegraphics[width=1\linewidth]{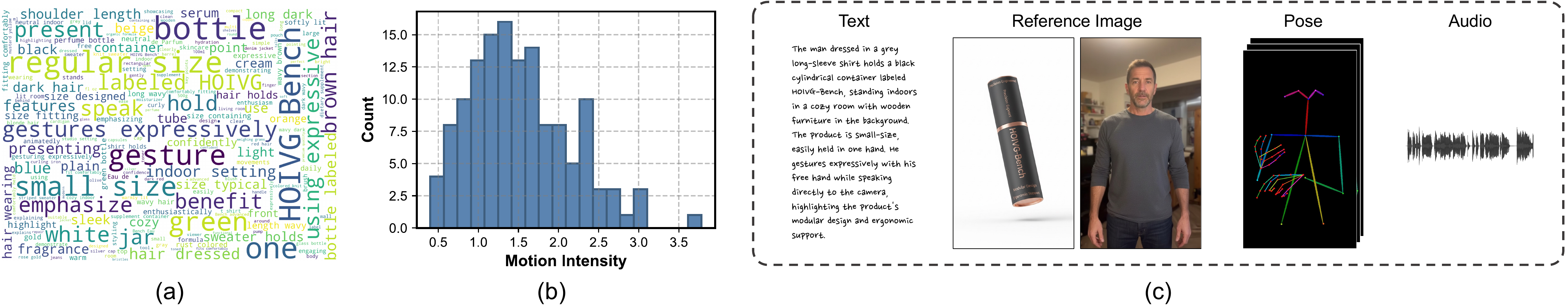}
    
    \caption
    { 
        \textbf{Statistics and example of HOIVG-Bench.}
        (a) Word cloud of text prompts illustrating the wide variety of human-object interaction scenarios. 
        (b) Motion intensity distribution of original videos calculated via dense optical flow \cite{farneback2003two}, reflecting a diverse dynamic range of pose conditions.
        (c) A representative sample from the benchmark, featuring text, reference images, pose, and audio.
        Constructed through a rigorous pipeline, the benchmark provides high-quality data to evaluate the synergy of diverse multimodal inputs.
    }
    
    \label{fig:bench_example}
    
\end{figure*}

Through the meticulous construction process described above, we have curated a high-quality benchmark, with statistics and an example visualized in \cref{fig:bench_example}. 
Note that the decision to use AI-generated human and object images was a carefully deliberated choice to comply with privacy, ethical, and legal guidelines for public release.
Moreover, to mitigate potential domain bias, we have conducted manual checks to filter out images with noticeable “AI-ness”, ensuring the benchmark closely reflects real-world data.

\myparagraph{Evaluation Metrics.}
The metrics are across five key dimensions:
(1) \textit{Text Alignment}: we employ VideoReward \cite{liu2025improving} to predict the text alignment (TA) score;
(2) \textit{Reference Consistency}: we follow OpenS2V \cite{yuan2025opens2v} to report FaceSim and NexusScore; 
(3) \textit{Pose Accuracy}: we use average keypoint distances (AKD) and percentage of correct keypoints (PCK) (with an error threshold of 5\%) based on DWPose \cite{yang2023effective};
(4) \textit{Audio-Visual Synchronization}: we adopt the Sync-C and Sync-D scores \cite{chung2016out}; and  
(5) \textit{Video Quality}: we use VBench \cite{huang2024vbench} for the AES and IQA metrics, plus VideoReward \cite{liu2025improving} for overall visual (VQ) and motion quality (MQ).
Although our model supports video generation of up to 10 seconds, to ensure a fair comparison with baseline methods that only support short-clip generation, all quantitative metrics and qualitative analyses on HOIVG-Bench are standardized to 5-second video clips at 720p resolution in portrait mode.

\begin{table*}[t]
     \caption{
     \textbf{Quantitative comparison.}
     We compare our \method with existing state-of-the-art methods, showing that \method can achieve superior/competitive performance across diverse multimodal conditioning settings.
     The best and second-best results in each setting are marked in \textbf{bold} and \underline{underlined}. The symbol ``-'' indicates that the metric is not applicable in the corresponding setting.
     }

    \centering
    \setlength{\tabcolsep}{2.5mm}
    \renewcommand{\arraystretch}{1.2}
    
    \resizebox{\linewidth}{!}
    {
\begin{tabular}{lccccccccccc}
\specialrule{0.1em}{0pt}{2pt}
\multirow{2}[4]{*}{Method} & Text Align. & \multicolumn{2}{c}{Reference Consistency} & \multicolumn{2}{c}{Audio-Visual Sync.} & \multicolumn{2}{c}{Pose Accuracy} & \multicolumn{4}{c}{Video Quality} \\

\cmidrule(lr){2-2} \cmidrule(lr){3-4} \cmidrule(lr){5-6} \cmidrule(lr){7-8} \cmidrule(lr){9-12}      & TA$\uparrow$ & FaceSim$\uparrow$ & NexusScore$\uparrow$ & Sync-C$\uparrow$ & Sync-D$\downarrow$ & AKD$\downarrow$ & PCK$\uparrow$ & AES$\uparrow$ & IQA$\uparrow$ & VQ$\uparrow$ & MQ$\uparrow$ \\
\midrule
\multicolumn{12}{c}{\textit{Text+Reference-to-Video (R2V)}} \\
\midrule
HunyuanCustom \cite{hu2025hunyuancustom} & 7.523 & 0.440 & 0.359 & -     & -     & -     & -     & 0.452 & 0.697 & 10.11 & 5.286 \\
HuMo-1.7B \cite{chen2025humo} & 7.087 & 0.647 & 0.333 & -     & -     & -     & -     & 0.441 & 0.723 & 9.76  & 3.406 \\
HuMo-17B \cite{chen2025humo} & 7.949 & 0.843 & 0.346 & -     & -     & -     & -     & 0.448 & 0.726 & 9.97  & 3.685 \\
VACE \cite{jiang2025vace} & \underline{8.413} & 0.759 & \underline{0.368} & -     & -     & -     & -     & 0.457 & 0.722 & 10.72  & 5.442 \\
Phantom-1.3B \cite{liu2025phantom} & 8.342 & 0.708 & 0.351 & -     & -     & -     & -     & \underline{0.459} & 0.722 & 10.90 & \underline{5.637} \\
Phantom-14B \cite{liu2025phantom} & \textbf{8.609} & \textbf{0.876} & 0.366 & -     & -     & -     & -     & 0.449 & \textbf{0.741} & \underline{10.93} & 5.517 \\
\method (Ours) & 7.746 & \underline{0.874} & \textbf{0.389} & -     & -     & -     & -     & \textbf{0.468} & \underline{0.740} & \textbf{11.12} & \textbf{5.885} \\
\midrule
\multicolumn{12}{c}{\textit{Text+Reference+Audio-to-Video (RA2V)}} \\
\midrule
HunyuanCustom \cite{hu2025hunyuancustom} & 7.289 & 0.457 & \underline{0.350} & 6.072 & 10.08 & -     & -     & \underline{0.439} & 0.715 & 9.15  & 3.658 \\
HuMo-1.7B \cite{chen2025humo} & 7.489 & 0.575 & 0.329 & 7.234 & 9.117 & -     & -     & 0.428 & 0.731 & 9.97  & 4.182 \\
HuMo-17B \cite{chen2025humo} & \textbf{8.146} & \underline{0.805} & 0.344 & \underline{8.013} & \underline{8.316} & -     & -     & 0.439 & \underline{0.739} & \underline{10.27} & \underline{4.269} \\
\method (Ours) & \underline{8.093} & \textbf{0.810} & \textbf{0.369} & \textbf{8.612} & \textbf{7.608} & -     & -     & \textbf{0.465} & \textbf{0.742} & \textbf{10.86} & \textbf{5.554} \\
\midrule
\multicolumn{12}{c}{\textit{Text+Reference+Pose-to-Video (RP2V)}} \\
\midrule
AnchorCrafter \cite{xu2024anchorcrafter} & 2.669 & 0.404 & 0.215 & -     & -     & 0.229 & 0.176 & \textbf{0.499} & 0.673 & 8.95  & 4.241 \\
VACE \cite{jiang2025vace} & \textbf{7.690} & \textbf{0.600} & \underline{0.352} & -     & -     & \underline{0.206} & \underline{0.336} & \underline{0.450} & \underline{0.712} & \underline{10.14} & \textbf{5.393} \\
\method (Ours) & \underline{6.526} & \underline{0.474} & \textbf{0.418} & -     & -     & \textbf{0.174} & \textbf{0.460} & 0.447 & \textbf{0.722} & \textbf{10.28} & \underline{4.937} \\
\specialrule{0.1em}{1pt}{0pt}
\end{tabular}

    } 
    \label{tab:quant_results}
\end{table*}

\section{Experiments}
\label{sec:exp}

\subsection{Setups}

\myparagraph{Implementation Details.}
We initialize our model based on Waver 1.0 \cite{zhang2025waver}. 
The overall training process spans two resolutions of 480p and 720p.
Unless otherwise specified, we conduct our large-scale training on a computing cluster of 128 GPUs, each with 80GB RAM. 
To ensure computational efficiency, we employ Fully Sharded Data Parallel (FSDP) \cite{zhao2023pytorch} combined with the Ulysses-style sequence parallelism \cite{jacobs2023deepspeed}, utilizing BF16 mixed precision. 
The model is optimized by AdamW \cite{loshchilov2017decoupled} with a learning rate of $3 \times 10^{-5}$ and a weight decay of $0.01$. 
More details are provided in \cref{sec:more_impl_details}.

\myparagraph{Compared Methods.}
We compare \method (12.3B) with leading open-source methods, including HunyuanCustom (13B) \cite{hu2025hunyuancustom}, HuMo (17B/1.7B) \cite{chen2025humo}, VACE (14B) \cite{jiang2025vace}, Phantom (14B/1.3B) \cite{liu2025phantom}, and AnchorCrafter (1.5B) \cite{xu2024anchorcrafter}. 
To ensure a thorough comparison, we evaluate several available variants with different model sizes for these baselines. 
However, as these methods do not support the full spectrum of four multimodal inputs inherent to \method, the experiments are conducted across varying input settings (\ie, R2V, RA2V, and RP2V) to align with the different capabilities of each baseline.

\subsection{Main Results}

\begin{figure*}[t]
    \centering
    
    \includegraphics[width=1\linewidth]{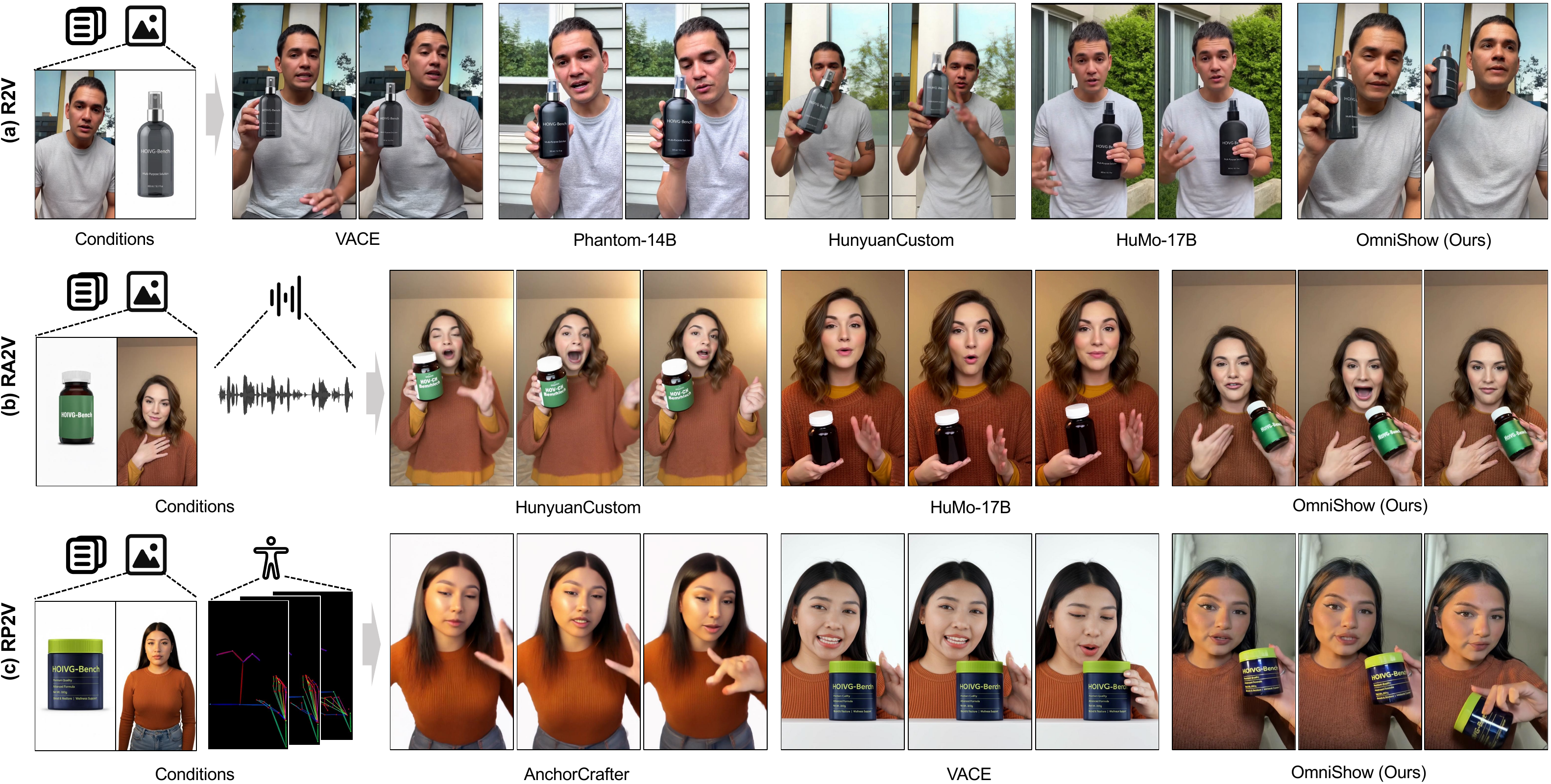}
    
    \caption
    { 
        \textbf{Qualitative comparison.}
       We present generated results from our \method and other methods across various multimodal condition settings. Note that text prompts are omitted for brevity.
    }
    
    \label{fig:qual_results}
    
\end{figure*}

\myparagraph{Quantitative Comparison.}
The quantitative results across three key settings are presented in \cref{tab:quant_results}. 
Our \method demonstrates robust performance, achieving state-of-the-art or highly competitive results across the majority of metrics.
Moreover, among 10B-scale models that meet industry-grade standards, OmniShow is the smallest and the most parameter-efficient.
In the R2V setting, our method matches the reference preservation capabilities of specialized methods like Phantom-14B, as evidenced by our comparable FaceSim and NexusScore.
Notably, \method exhibits distinct advantages as a unified framework in more complex scenarios. 
In the RA2V setting, while dedicated baselines like HuMo-17B may show slight gains in TA, our approach delivers leading performance in other metrics like NexusScore and Sync-C. 
Moving to the RP2V setting, although the precise pose adherence introduces viewpoint shifts and facial morphology changes that affect FaceSim, \method can still maintain robust video quality.
These results underscore its ability to effectively orchestrate multimodal conditions within a cohesive architecture.
It is worth noting that \method is the only model supporting RAP2V generation, and there are no directly comparable counterparts. 
To further verify its capabilities, we compare \method with a cascaded baseline in \cref{sec:more_comp}.

\myparagraph{Qualitative Comparison.}
The qualitative results in \cref{fig:qual_results} highlight the superiority of our method.
Across varied scenarios, \method consistently exhibits high-fidelity reference preservation, natural motion dynamics, and precise audio-visual synchronization.
In the R2V setting, unlike baselines that often rigidly paste objects onto human subjects at implausible sizes, our model ensures both visual fidelity and realistic composition.
For the RA2V setting, \method generates natural body movements alongside precise lip synchronization, avoiding the ``overreaction'' and ``frozen body'' issue common in previous methods.
Crucially, in the RP2V setting involving complex motion, our model demonstrates robustness in handling complex spatial interactions and large pose variations, accurately generating hand contacts and object appearance, whereas VACE struggles with following the pose and AnchorCrafter fails to preserve object identity.
Since no existing methods simultaneously support all four conditions as ours do, a direct comparison in this full setting (\ie, RAP2V) is absent. 
Moreover, we showcase more qualitative results across all four multimodal conditioning settings in \cref{fig:teaser} and \cref{sec:more_qual_results}, which further verify its effectiveness in unifying all modalities simultaneously to generate coherent, expressive, and realistic videos.

\begin{figure}[t]

    \centering

    \includegraphics[width=1\linewidth]{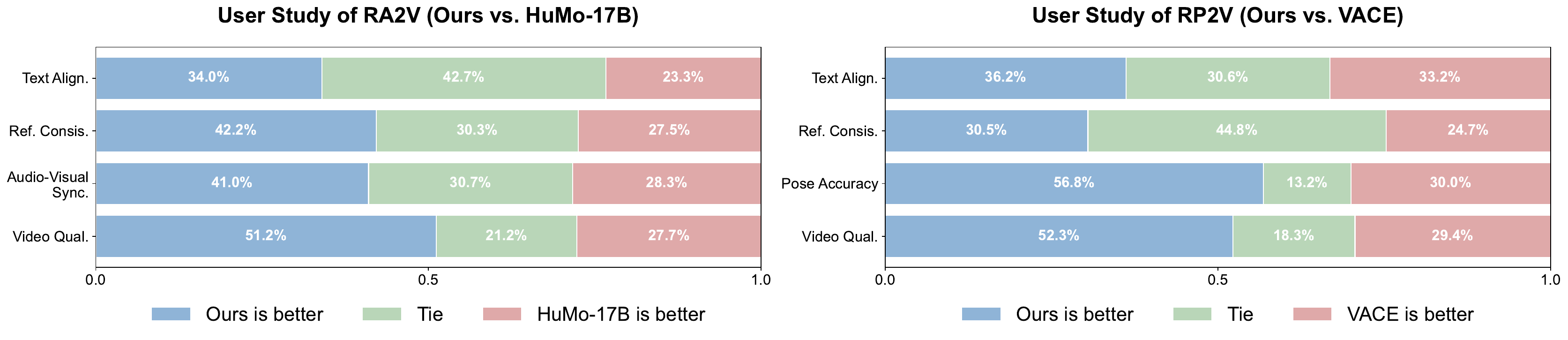}

    \caption
    { 
       \textbf{Side-by-side human evaluation.} Judged by human evaluators, our \method achieves superior performance in overall quality, showing great alignment with human preference.    
    }

    \label{fig:sbs}

\end{figure}

\myparagraph{Human Evaluation.}
To measure human preference, we conduct side-by-side comparisons in both the RA2V and RP2V settings, where user experience is paramount.
Specifically, we engage a diverse demographic pool of evaluators, including 30 participants for the RA2V task and 33 participants for the RP2V task, to assess randomly selected subsets of 20 samples.
As shown in \cref{fig:sbs}, the user studies show that human evaluators favor our results even when some objective scores are comparable. 
Moreover, our \method is preferred in the majority of cases, demonstrating superiority in both condition adherence and overall visual quality.
We attribute this to smoother temporal dynamics and richer visual details in our generated videos, which are critical for perceived realism but might be overlooked by frame-level metrics.
These evaluation results further highlight the superiority of our \method.

\subsection{Ablation Studies and Analysis}

In this section, we present a series of experiments to further evaluate the effectiveness of \method, which include 
(i) comprehensive ablation studies on the proposed techniques, 
(ii) an additional evaluation on the A2V task, and 
(iii) an exploration of our model's broader applications. 
Note that unless otherwise stated, all ablation training runs, including those for our methods, are conducted on 8 GPUs and ensure fairness.

\myparagraph{Ablation of \techVision.}
We first compare this mechanism against the token concatenation approach \cite{tan2025ominicontrol} in the R2V task. 
Results in \cref{tab:abl_tech_vision} show our method yields superior video quality and reference consistency (also see \cref{fig:channel_concat}). 
This highlights that adhering to the native conditioning structure is more efficient for visual injection in task-unified video generation models.
Additionally, removing the reference reconstruction loss leads to a drop in visual fidelity, especially for human identity, validating its role in enforcing semantic preservation.

\myparagraph{Ablation of \techAudio.}
We evaluate variants by removing audio context, attention map constraints, or adaptive gating in the A2V task. 
Results in \cref{tab:abl_tech_audio} indicate that incorporating audio context is crucial for capturing temporal coherence, reflected in improved Sync-D, while attention map constraints significantly boosts synchronization by ensuring frame-wise interaction. 
Furthermore, disabling adaptive gating degrades final visual quality, confirming the necessity of audio feature modulation for training stability.

\myparagraph{Ablation of \techTrain.}
We compare our training strategy against conventional ones using 32 GPUs, as shown in \cref{tab:abl_tech_train}.
Specifically, ``Only RA2V'' means training directly on RA2V data, while ``R2V/A2V$\to$RA2V'' denotes training on R2V/A2V data and then RA2V data.
These baselines fail to fully leverage heterogeneous data: single-stage training underperforms due to poor convergence, while naive multi-stage approaches struggle with incorporating new modal inputs. 
In contrast, our paradigm achieves the best trade-off between reference consistency and audio-visual synchronization.

\myparagraph{Effectiveness on the A2V Task.}
As audio is arguably the most special modality to tame, we conduct an additional evaluation on the model derived from the A2V training, denoted as \textsc{OmniShow-A2V}.
Benefiting from our \techAudio, this model achieves state-of-the-art performance on the EMTD benchmark \cite{meng2025echomimicv2}, as illustrated in \cref{tab:a2v}.
Notably, its Sync-C score of 6.49 surpasses competitive methods like MultiTalk \cite{kong2025let}.
These results validate the efficacy of the proposed technique on the A2V task, which also establishes a robust foundation for the final unified model.

\myparagraph{Broader Applications.}
Beyond standard benchmarks, \method exhibits remarkable versatility in broader applications, as illustrated in \cref{fig:teaser}b.
First, it naturally supports \textit{audio-driven avatars} as a subset capability. Given a human reference image and an audio input, it effectively animates characters with synchronized speech and natural expressions.
Moreover, \method enables more complex creative tasks by integrating into specific workflows.
For instance, in \textit{object swapping}, we can replace the original object in a video with a novel one while keeping the human subject and pose consistent.
Furthermore, \textit{video remixing} allows for synthesizing new videos by recombining pose, object, and human references from different sources.
These results highlight its strong compositional control, paving the way for diverse real-world content creation.
\section{Discussion}

\myparagraph{Insights.}
Beyond the empirical results, we highlight two key insights derived from our work: 
(1) \textit{Philosophy of Minimalist Intervention}: 
We demonstrated that deeply understanding the underlying input structure and learning dynamics of DiTs enables the design of efficient injection methods. 
Specifically, we repurpose the native channel-concat mechanism and introduce attention modules via the analysis of gate vectors, ultimately empowering superior multimodal control with minimal architectural changes.
(2) \textit{Evolution from Specialists to a Generalist}:  
We showed that rather than relying solely on scarce paired data, one can fully leverage heterogeneous data through a decoupled-then-joint training paradigm. Crucially, we found that striking RA2V capabilities can emerge simply by merging R2V and A2V models. This finding, that controllability can emerge via weight merging, is also an inspiring discovery that we hope will motivate further exploration.

\myparagraph{Limitations.}
Although \method achieves promising results, it still has certain limitations. 
First, while our model is capable of generating longer videos (up to 10 seconds), our current evaluation focuses on 5-second clips to ensure a fair comparison with baselines that only support short-clip generation. 
Second, the human reference images in our benchmark is AI-generated, which might introduce slight distribution biases compared to purely real-world photos. 
Finally, in some extreme scenarios involving overly intense motion or conflicting multimodal inputs, the model may occasionally exhibit artifacts or blur in generated videos.

\myparagraph{Future Work.}
Looking ahead, several promising directions could further elevate the capabilities of \method. 
First, Reinforcement Learning (RL)-based post-training methods are worth fully exploring in this domain. By incorporating rewards tailored to human preference and physical plausibility, we aim to better align the generation with real-world dynamics and mitigate artifacts under extreme conditions.
Moreover, we aim to scale up the training data and model capacity to push the boundaries of the model's generalization ability in complex scenarios. 
Ultimately, we envision expanding the framework to support richer multimodal inputs (\eg, camera trajectories or reference videos) and exploring broader model capabilities, such as agentic minute-level video synthesis and streaming interactive generation.

\begin{table*}[t]
    \caption{\textbf{Quantitative results of ablation studies.} We evaluate three key components of \method, including (a) \techVision, (b) \techAudio,  and (c) \techTrain.}
    \label{tab:full_ablation}
    
    \centering
    \renewcommand{\arraystretch}{1.1}

    \begin{subtable}{0.31\linewidth}
        \begin{minipage}[t]{0.08\linewidth}
            \vspace{0pt} 
            {\scriptsize (a)} 
        \end{minipage}
        \begin{minipage}[t]{0.92\linewidth}
            \vspace{0pt} 
            \setlength{\tabcolsep}{0.8mm} 
            \resizebox{\linewidth}{!}{
                \begin{tabular}[t]{lccc} 
                \toprule
                Variant & FaceSim$\uparrow$ & NexusScore$\uparrow$ & AES$\uparrow$ \\
                \midrule
                Token-Concat & 0.601 & 0.344 & \underline{0.466} \\
                w/o Ref. Rec. Loss & \underline{0.678} & \underline{0.352} & 0.466 \\
                Ours  & \textbf{0.707} & \textbf{0.353} & \textbf{0.471} \\
                \bottomrule
                \end{tabular}
            }
        \end{minipage}
        \phantomcaption\label{tab:abl_tech_vision}
    \end{subtable}
    \hfill
    \begin{subtable}{0.322\linewidth}
        \begin{minipage}[t]{0.075\linewidth}
            \vspace{0pt}
            {\scriptsize (b)}
        \end{minipage}
        \begin{minipage}[t]{0.925\linewidth}
            \vspace{0pt}
            \setlength{\tabcolsep}{0.8mm}
            \resizebox{\linewidth}{!}{
                \begin{tabular}[t]{lccc}
                \toprule
                Variant & Sync-C$\uparrow$ & Sync-D$\downarrow$ & AES$\uparrow$ \\
                \midrule
                w/o Audio Context & \underline{8.872} & 7.878 & 0.533 \\
                w/o Attn.-Map Constraints & 2.201 & 13.01 & \textbf{0.545} \\
                w/o Adaptive Gating & \underline{8.872} & \underline{7.819} & 0.529 \\
                Ours  & \textbf{9.023} & \textbf{7.419} & \underline{0.540} \\
                \bottomrule
                \end{tabular}
            }
        \end{minipage}
        \phantomcaption\label{tab:abl_tech_audio}
    \end{subtable}
    \hfill
    \begin{subtable}{0.353\linewidth}
        \begin{minipage}[t]{0.065\linewidth}
            \vspace{0pt}
            {\scriptsize (c)}
        \end{minipage}
        \begin{minipage}[t]{0.935\linewidth}
            \vspace{0pt}
            \setlength{\tabcolsep}{0.8mm}
            \resizebox{\linewidth}{!}{
                \begin{tabular}[t]{lccc}
                \toprule
                Variant & NexusScore$\uparrow$ & Sync-D$\downarrow$ & AES$\uparrow$ \\
                \midrule
                Single-Stage (Only RA2V) & 0.345 & 13.11 & 0.453 \\
                Multi-Stage (R2V $\rightarrow$ RA2V) & \underline{0.360} & 13.23 & \underline{0.473} \\
                Multi-Stage (A2V $\rightarrow$ RA2V) & 0.342 & \textbf{7.38} & 0.456 \\
                Ours  & \textbf{0.364} & \underline{8.14} & \textbf{0.474} \\
                \bottomrule
                \end{tabular}
            }
        \end{minipage}
        \phantomcaption\label{tab:abl_tech_train}
    \end{subtable}
    
\end{table*}
\begin{table}[t]
     \caption{
     \textbf{Performance on the EMTD benchmark.}
     Note that metrics marked with “*” adopt the definitions from this benchmark, thus leading to different numerical ranges.
     }

    \centering
    \setlength{\tabcolsep}{6mm}
    \renewcommand{\arraystretch}{1.2}
    
    \resizebox{0.7\linewidth}{!}
    {
\begin{tabular}{lcccc}
\toprule
Method & *IQA$\uparrow$ & *AES$\uparrow$ & Sync-C$\uparrow$ & Sync-D$\downarrow$ \\
\midrule
FantasyTalking \cite{wang2025fantasytalking} & 2.11  & 1.12  & 1.11  & 12.88 \\
HunyuanVideo-Avatar \cite{chen2025hunyuanvideo} & 1.76  & 1.18  & 4.89  & 9.37 \\
Hallo3 \cite{cui2025hallo3} & \textbf{2.31}  & \underline{1.48}  & 4.26  & 10.22 \\
MultiTalk \cite{kong2025let} & 2.07  & 1.30  & \underline{6.34}  & \textbf{8.47} \\
OmniAvatar \cite{gan2025omniavatar} & 2.16  & 1.31  & 5.40  & 9.13 \\
\textsc{OmniShow-A2V} (Ours) & \underline{2.26} & \textbf{1.51} & \textbf{6.49} & \underline{8.97} \\
\bottomrule
\end{tabular}
    }    
    \vspace{-2mm}
    \label{tab:a2v}
\end{table}

\section{Conclusion}
\label{sec:concl}

In this work, we introduced \textbf{\method}, a unified framework for HOIVG. 
By orchestrating text, reference image, audio, and pose conditions, \method achieves precise multimodal control and high-quality video generation. 
Through the carefully designed architecture and training strategy, we effectively achieve harmonious multimodal injection and heterogeneous data utility. 
Extensive comparison and detailed ablation studies on HOIVG-Bench verified the superiority of \method. 
In the future, we envision expanding its capabilities with larger datasets, richer inputs, and broader fields.

\section*{Impact Statement}

While our primary goal is to push the boundaries of video generation research, we acknowledge the potential societal implications, including both positive impacts, such as enhancing accessibility in digital content creation and education, and risks, such as misuse for generating deceptive or harmful content.
To proactively mitigate these concerns, we advocate for responsible use and the continuous development of safeguards to ensure this technology ultimately serves constructive and beneficial societal purposes.

\section*{Acknowledgments}
We thank Qijun Gan, Pan Xie, and Yifu Zhang for helpful discussions, and Liyang Chen for his advice on baseline reproduction. We appreciate the support from Yuqi Zhang and Hao Yang regarding internal devkits, and the assistance from Ruibiao Lu, Chao Zhang, and Wei Feng on data collection.

\clearpage

\bibliographystyle{plainnat}
\bibliography{main}

@article{zhao2023pytorch,
  title={Pytorch fsdp: experiences on scaling fully sharded data parallel},
  author={Zhao, Yanli and Gu, Andrew and Varma, Rohan and Luo, Liang and Huang, Chien-Chin and Xu, Min and Wright, Less and Shojanazeri, Hamid and Ott, Myle and Shleifer, Sam and others},
  journal={arXiv preprint arXiv:2304.11277},
  year={2023}
}

@article{zhang2025waver,
  title={Waver: Wave your way to lifelike video generation},
  author={Zhang, Yifu and Yang, Hao and Zhang, Yuqi and Hu, Yifei and Zhu, Fengda and Lin, Chuang and Mei, Xiaofeng and Jiang, Yi and Peng, Bingyue and Yuan, Zehuan},
  journal={arXiv preprint arXiv:2508.15761},
  year={2025}
}

@article{jacobs2023deepspeed,
  title={Deepspeed ulysses: System optimizations for enabling training of extreme long sequence transformer models},
  author={Jacobs, Sam Ade and Tanaka, Masahiro and Zhang, Chengming and Zhang, Minjia and Song, Shuaiwen Leon and Rajbhandari, Samyam and He, Yuxiong},
  journal={arXiv preprint arXiv:2309.14509},
  year={2023}
}

@article{wan2025wan,
  title={Wan: Open and advanced large-scale video generative models},
  author={Wan, Team and Wang, Ang and Ai, Baole and Wen, Bin and Mao, Chaojie and Xie, Chen-Wei and Chen, Di and Yu, Feiwu and Zhao, Haiming and Yang, Jianxiao and others},
  journal={arXiv preprint arXiv:2503.20314},
  year={2025}
}

@article{chen2025seedance,
  title={Seedance 1.5 pro: A Native Audio-Visual Joint Generation Foundation Model},
  author={Chen, Siyan and Chen, Yanfei and Chen, Ying and Chen, Zhuo and Cheng, Feng and Chi, Xuyan and Cong, Jian and Cui, Qinpeng and Dong, Qide and Fan, Junliang and others},
  journal={arXiv preprint arXiv:2512.13507},
  year={2025}
}

@article{wu2025hunyuanvideo,
  title={HunyuanVideo 1.5 Technical Report},
  author={Wu, Bing and Zou, Chang and Li, Changlin and Huang, Duojun and Yang, Fang and Tan, Hao and Peng, Jack and Wu, Jianbing and Xiong, Jiangfeng and Jiang, Jie and others},
  journal={arXiv preprint arXiv:2511.18870},
  year={2025}
}

@article{liu2025phantom,
  title={Phantom: Subject-consistent video generation via cross-modal alignment},
  author={Liu, Lijie and Ma, Tianxiang and Li, Bingchuan and Chen, Zhuowei and Liu, Jiawei and Li, Gen and Zhou, Siyu and He, Qian and Wu, Xinglong},
  journal={arXiv preprint arXiv:2502.11079},
  year={2025}
}

@article{li2024latentsync,
  title={Latentsync: Taming audio-conditioned latent diffusion models for lip sync with syncnet supervision},
  author={Li, Chunyu and Zhang, Chao and Xu, Weikai and Lin, Jingyu and Xie, Jinghui and Feng, Weiguo and Peng, Bingyue and Chen, Cunjian and Xing, Weiwei},
  journal={arXiv preprint arXiv:2412.09262},
  year={2024}
}

@article{fei2025skyreels,
  title={Skyreels-a2: Compose anything in video diffusion transformers},
  author={Fei, Zhengcong and Li, Debang and Qiu, Di and Wang, Jiahua and Dou, Yikun and Wang, Rui and Xu, Jingtao and Fan, Mingyuan and Chen, Guibin and Li, Yang and others},
  journal={arXiv preprint arXiv:2504.02436},
  year={2025}
}

@article{gan2025omniavatar,
  title={OmniAvatar: Efficient Audio-Driven Avatar Video Generation with Adaptive Body Animation},
  author={Gan, Qijun and Yang, Ruizi and Zhu, Jianke and Xue, Shaofei and Hoi, Steven},
  journal={arXiv preprint arXiv:2506.18866},
  year={2025}
}

@article{hacohen2026ltx,
  title={LTX-2: Efficient Joint Audio-Visual Foundation Model},
  author={HaCohen, Yoav and Brazowski, Benny and Chiprut, Nisan and Bitterman, Yaki and Kvochko, Andrew and Berkowitz, Avishai and Shalem, Daniel and Lifschitz, Daphna and Moshe, Dudu and Porat, Eitan and others},
  journal={arXiv preprint arXiv:2601.03233},
  year={2026}
}

@article{hu2025hunyuancustom,
  title={Hunyuancustom: A multimodal-driven architecture for customized video generation},
  author={Hu, Teng and Yu, Zhentao and Zhou, Zhengguang and Liang, Sen and Zhou, Yuan and Lin, Qin and Lu, Qinglin},
  journal={arXiv preprint arXiv:2505.04512},
  year={2025}
}

@inproceedings{farneback2003two,
  title={Two-frame motion estimation based on polynomial expansion},
  author={Farneb{\"a}ck, Gunnar},
  booktitle={Scandinavian conference on Image analysis},
  pages={363--370},
  year={2003},
  organization={Springer}
}

@article{chen2025humo,
  title={Humo: Human-centric video generation via collaborative multi-modal conditioning},
  author={Chen, Liyang and Ma, Tianxiang and Liu, Jiawei and Li, Bingchuan and Chen, Zhuowei and Liu, Lijie and He, Xu and Li, Gen and He, Qian and Wu, Zhiyong},
  journal={arXiv preprint arXiv:2509.08519},
  year={2025}
}

@article{wang2025dreamactor,
  title={Dreamactor-h1: High-fidelity human-product demonstration video generation via motion-designed diffusion transformers},
  author={Wang, Lizhen and Xia, Zhurong and Hu, Tianshu and Wang, Pengrui and Wei, Pengfei and Zheng, Zerong and Zhou, Ming and Zhang, Yuan and Gao, Mingyuan},
  journal={arXiv preprint arXiv:2506.10568},
  year={2025}
}

@article{huang2025jova,
  title={JoVA: Unified Multimodal Learning for Joint Video-Audio Generation},
  author={Huang, Xiaohu and Zhou, Hao and Yang, Qiangpeng and Wen, Shilei and Han, Kai},
  journal={arXiv preprint arXiv:2512.13677},
  year={2025}
}

@inproceedings{meng2025echomimicv2,
  title={Echomimicv2: Towards striking, simplified, and semi-body human animation},
  author={Meng, Rang and Zhang, Xingyu and Li, Yuming and Ma, Chenguang},
  booktitle={Proceedings of the Computer Vision and Pattern Recognition Conference},
  pages={5489--5498},
  year={2025}
}

@article{low2025ovi,
  title={Ovi: Twin backbone cross-modal fusion for audio-video generation},
  author={Low, Chetwin and Wang, Weimin and Katyal, Calder},
  journal={arXiv preprint arXiv:2510.01284},
  year={2025}
}

@inproceedings{cui2025hallo3,
  title={Hallo3: Highly dynamic and realistic portrait image animation with video diffusion transformer},
  author={Cui, Jiahao and Li, Hui and Zhan, Yun and Shang, Hanlin and Cheng, Kaihui and Ma, Yuqi and Mu, Shan and Zhou, Hang and Wang, Jingdong and Zhu, Siyu},
  booktitle={Proceedings of the Computer Vision and Pattern Recognition Conference},
  pages={21086--21095},
  year={2025}
}

@inproceedings{yuan2025identity,
  title={Identity-preserving text-to-video generation by frequency decomposition},
  author={Yuan, Shenghai and Huang, Jinfa and He, Xianyi and Ge, Yunyang and Shi, Yujun and Chen, Liuhan and Luo, Jiebo and Yuan, Li},
  booktitle={Proceedings of the Computer Vision and Pattern Recognition Conference},
  pages={12978--12988},
  year={2025}
}

@article{zhou2025scaling,
  title={Scaling Zero-Shot Reference-to-Video Generation},
  author={Zhou, Zijian and Liu, Shikun and Liu, Haozhe and Qiu, Haonan and An, Zhaochong and Ren, Weiming and Liu, Zhiheng and Huang, Xiaoke and Ng, Kam Woh and Xie, Tian and others},
  journal={arXiv preprint arXiv:2512.06905},
  year={2025}
}

@inproceedings{zhang2023sadtalker,
  title={Sadtalker: Learning realistic 3d motion coefficients for stylized audio-driven single image talking face animation},
  author={Zhang, Wenxuan and Cun, Xiaodong and Wang, Xuan and Zhang, Yong and Shen, Xi and Guo, Yu and Shan, Ying and Wang, Fei},
  booktitle={Proceedings of the IEEE/CVF conference on computer vision and pattern recognition},
  pages={8652--8661},
  year={2023}
}

@article{jiang2024loopy,
  title={Loopy: Taming audio-driven portrait avatar with long-term motion dependency},
  author={Jiang, Jianwen and Liang, Chao and Yang, Jiaqi and Lin, Gaojie and Zhong, Tianyun and Zheng, Yanbo},
  journal={arXiv preprint arXiv:2409.02634},
  year={2024}
}

@article{cui2025hallo4,
  title={Hallo4: High-Fidelity Dynamic Portrait Animation via Direct Preference Optimization and Temporal Motion Modulation},
  author={Cui, Jiahao and Chen, Yan and Xu, Mingwang and Shang, Hanlin and Chen, Yuxuan and Zhan, Yun and Dong, Zilong and Yao, Yao and Wang, Jingdong and Zhu, Siyu},
  journal={arXiv preprint arXiv:2505.23525},
  year={2025}
}

@inproceedings{lin2025omnihuman,
  title={Omnihuman-1: Rethinking the scaling-up of one-stage conditioned human animation models},
  author={Lin, Gaojie and Jiang, Jianwen and Yang, Jiaqi and Zheng, Zerong and Liang, Chao and Zhang, Yuan and Liu, Jingtuo},
  booktitle={Proceedings of the IEEE/CVF International Conference on Computer Vision},
  pages={13847--13858},
  year={2025}
}

@article{yi2025magicinfinite,
  title={Magicinfinite: Generating infinite talking videos with your words and voice},
  author={Yi, Hongwei and Ye, Tian and Shao, Shitong and Yang, Xuancheng and Zhao, Jiantong and Guo, Hanzhong and Wang, Terrance and Yin, Qingyu and Xie, Zeke and Zhu, Lei and others},
  journal={arXiv preprint arXiv:2503.05978},
  year={2025}
}

@article{kong2025let,
  title={Let Them Talk: Audio-Driven Multi-Person Conversational Video Generation},
  author={Kong, Zhe and Gao, Feng and Zhang, Yong and Kang, Zhuoliang and Wei, Xiaoming and Cai, Xunliang and Chen, Guanying and Luo, Wenhan},
  journal={arXiv preprint arXiv:2505.22647},
  year={2025}
}

@article{wang2025interacthuman,
  title={InterActHuman: Multi-Concept Human Animation with Layout-Aligned Audio Conditions},
  author={Wang, Zhenzhi and Yang, Jiaqi and Jiang, Jianwen and Liang, Chao and Lin, Gaojie and Zheng, Zerong and Yang, Ceyuan and Lin, Dahua},
  journal={arXiv preprint arXiv:2506.09984},
  year={2025}
}

@article{team2025kling,
  title={Kling-Omni Technical Report},
  author={{Kling Team} and Chen, Jialu and Ci, Yuanzheng and Du, Xiangyu and Feng, Zipeng and Gai, Kun and Guo, Sainan and Han, Feng and He, Jingbin and He, Kang and others},
  journal={arXiv preprint arXiv:2512.16776},
  year={2025}
}

@article{zhong2025anytalker,
  title={AnyTalker: Scaling Multi-Person Talking Video Generation with Interactivity Refinement},
  author={Zhong, Zhizhou and Ji, Yicheng and Kong, Zhe and Liu, Yiying and Wang, Jiarui and Feng, Jiasun and Liu, Lupeng and Wang, Xiangyi and Li, Yanjia and She, Yuqing and others},
  journal={arXiv preprint arXiv:2511.23475},
  year={2025}
}

@article{xue2024follow,
  title={Follow-your-pose v2: Multiple-condition guided character image animation for stable pose control},
  author={Xue, Jingyun and Wang, Hongfa and Tian, Qi and Ma, Yue and Wang, Andong and Zhao, Zhiyuan and Min, Shaobo and Zhao, Wenzhe and Zhang, Kaihao and Shum, Heung-Yeung and others},
  journal={arXiv e-prints},
  pages={arXiv--2406},
  year={2024}
}

@article{gan2025humandit,
  title={Humandit: Pose-guided diffusion transformer for long-form human motion video generation},
  author={Gan, Qijun and Ren, Yi and Zhang, Chen and Ye, Zhenhui and Xie, Pan and Yin, Xiang and Yuan, Zehuan and Peng, Bingyue and Zhu, Jianke},
  journal={arXiv preprint arXiv:2502.04847},
  year={2025}
}

@inproceedings{xu2024magicanimate,
  title={Magicanimate: Temporally consistent human image animation using diffusion model},
  author={Xu, Zhongcong and Zhang, Jianfeng and Liew, Jun Hao and Yan, Hanshu and Liu, Jia-Wei and Zhang, Chenxu and Feng, Jiashi and Shou, Mike Zheng},
  booktitle={Proceedings of the IEEE/CVF Conference on Computer Vision and Pattern Recognition},
  pages={1481--1490},
  year={2024}
}

@article{jiang2025vace,
  title={VACE: All-in-one video creation and editing},
  author={Jiang, Zeyinzi and Han, Zhen and Mao, Chaojie and Zhang, Jingfeng and Pan, Yulin and Liu, Yu},
  journal={arXiv preprint arXiv:2503.07598},
  year={2025}
}

@inproceedings{diller2024cg,
  title={Cg-hoi: Contact-guided 3d human-object interaction generation},
  author={Diller, Christian and Dai, Angela},
  booktitle={Proceedings of the IEEE/CVF Conference on Computer Vision and Pattern Recognition},
  pages={19888--19901},
  year={2024}
}

@inproceedings{peng2025hoi,
  title={Hoi-diff: Text-driven synthesis of 3d human-object interactions using diffusion models},
  author={Peng, Xiaogang and Xie, Yiming and Wu, Zizhao and Jampani, Varun and Sun, Deqing and Jiang, Huaizu},
  booktitle={Proceedings of the Computer Vision and Pattern Recognition Conference},
  pages={2878--2888},
  year={2025}
}

@article{shao2025magicdistillation,
  title={Magicdistillation: Weak-to-strong video distillation for large-scale few-step synthesis},
  author={Shao, Shitong and Yi, Hongwei and Guo, Hanzhong and Ye, Tian and Zhou, Daquan and Lingelbach, Michael and Xu, Zhiqiang and Xie, Zeke},
  journal={arXiv preprint arXiv:2503.13319},
  year={2025}
}

@article{xu2021d3d,
  title={D3d-hoi: Dynamic 3d human-object interactions from videos},
  author={Xu, Xiang and Joo, Hanbyul and Mori, Greg and Savva, Manolis},
  journal={arXiv preprint arXiv:2108.08420},
  year={2021}
}

@article{wang2025wisa,
  title={Wisa: World simulator assistant for physics-aware text-to-video generation},
  author={Wang, Jing and Ma, Ao and Cao, Ke and Zheng, Jun and Zhang, Zhanjie and Feng, Jiasong and Liu, Shanyuan and Ma, Yuhang and Cheng, Bo and Leng, Dawei and others},
  journal={arXiv preprint arXiv:2503.08153},
  year={2025}
}

@article{qin2026innoads,
  title={InnoAds-Composer: Efficient Condition Composition for E-Commerce Poster Generation},
  author={Qin, Yuxin and Cao, Ke and Liu, Haowei and Ma, Ao and Li, Fengheng and Zhu, Honghe and Zhang, Zheng and Ling, Run and Feng, Wei and He, Xuanhua and others},
  journal={arXiv preprint arXiv:2603.05898},
  year={2026}
}

@inproceedings{ling2026mofu,
  title={Mofu: Scale-aware modulation and fourier fusion for multi-subject video generation},
  author={Ling, Run and Cao, Ke and Lu, Jian and Ma, Ao and Liu, Haowei and He, Runze and Wang, Changwei and Xu, Rongtao and Shao, Yihua and Zhang, Zhanjie and others},
  booktitle={Proceedings of the AAAI Conference on Artificial Intelligence},
  volume={40},
  number={9},
  pages={7033--7041},
  year={2026}
}

@inproceedings{ye2023diffusion,
  title={Diffusion-guided reconstruction of everyday hand-object interaction clips},
  author={Ye, Yufei and Hebbar, Poorvi and Gupta, Abhinav and Tulsiani, Shubham},
  booktitle={Proceedings of the IEEE/CVF international conference on computer vision},
  pages={19717--19728},
  year={2023}
}

@InProceedings{Fan_2025_CVPR,
    author    = {Fan, Yingying and Yang, Quanwei and Wang, Kaisiyuan and Zhou, Hang and Li, Yingying and Feng, Haocheng and Ding, Errui and Wu, Yu and Wang, Jingdong},
    title     = {Re-HOLD: Video Hand Object Interaction Reenactment via adaptive Layout-instructed Diffusion Model},
    booktitle = {Proceedings of the IEEE/CVF Conference on Computer Vision and Pattern Recognition (CVPR)},
    month     = {June},
    year      = {2025},
    pages     = {17550-17560}
}

@article{xue2024hoi,
  title={Hoi-swap: Swapping objects in videos with hand-object interaction awareness},
  author={Xue, Zihui Sherry and Luo, Romy and Chen, Changan and Grauman, Kristen},
  journal={Advances in Neural Information Processing Systems},
  volume={37},
  pages={77132--77164},
  year={2024}
}

@article{chen2026posteromni,
  title={PosterOmni: Generalized Artistic Poster Creation via Task Distillation and Unified Reward Feedback},
  author={Chen, Sixiang and Lai, Jianyu and Gao, Jialin and Shi, Hengyu and Liu, Zhongying and Ye, Tian and Luo, Junfeng and Wei, Xiaoming and Zhu, Lei},
  journal={arXiv preprint arXiv:2602.12127},
  year={2026}
}

@article{chen2024virtualmodel,
  title={Virtualmodel: Generating object-id-retentive human-object interaction image by diffusion model for e-commerce marketing},
  author={Chen, Binghui and Zhong, Chongyang and Xiang, Wangmeng and Geng, Yifeng and Xie, Xuansong},
  journal={arXiv preprint arXiv:2405.09985},
  year={2024}
}

@article{xu2024anchorcrafter,
  title={Anchorcrafter: Animate cyberanchors saling your products via human-object interacting video generation},
  author={Xu, Ziyi and Huang, Ziyao and Cao, Juan and Zhang, Yong and Cun, Xiaodong and Shuai, Qing and Wang, Yuchen and Bao, Linchao and Li, Jintao and Tang, Fan},
  journal={arXiv preprint arXiv:2411.17383},
  year={2024}
}

@article{huang2025hunyuanvideo,
  title={HunyuanVideo-HOMA: Generic Human-Object Interaction in Multimodal Driven Human Animation},
  author={Huang, Ziyao and Zhou, Zixiang and Cao, Juan and Ma, Yifeng and Chen, Yi and Rao, Zejing and Xu, Zhiyong and Wang, Hongmei and Lin, Qin and Zhou, Yuan and others},
  journal={arXiv preprint arXiv:2506.08797},
  year={2025}
}

@article{chen2025hunyuanvideo,
  title={HunyuanVideo-Avatar: High-Fidelity Audio-Driven Human Animation for Multiple Characters},
  author={Chen, Yi and Liang, Sen and Zhou, Zixiang and Huang, Ziyao and Ma, Yifeng and Tang, Junshu and Lin, Qin and Zhou, Yuan and Lu, Qinglin},
  journal={arXiv preprint arXiv:2505.20156},
  year={2025}
}

@inproceedings{wang2025fantasytalking,
  title={Fantasytalking: Realistic talking portrait generation via coherent motion synthesis},
  author={Wang, Mengchao and Wang, Qiang and Jiang, Fan and Fan, Yaqi and Zhang, Yunpeng and Qi, Yonggang and Zhao, Kun and Xu, Mu},
  booktitle={Proceedings of the 33rd ACM International Conference on Multimedia},
  pages={9891--9900},
  year={2025}
}

@article{lin2025jarvisevo,
  title={JarvisEvo: Towards a Self-Evolving Photo Editing Agent with Synergistic Editor-Evaluator Optimization},
  author={Lin, Yunlong and Wang, Linqing and Lin, Kunjie and Lin, Zixu and Gong, Kaixiong and Li, Wenbo and Lin, Bin and Li, Zhenxi and Zhang, Shiyi and Peng, Yuyang and others},
  journal={arXiv preprint arXiv:2511.23002},
  year={2025}
}

@article{lipman2022flow,
  title={Flow matching for generative modeling},
  author={Lipman, Yaron and Chen, Ricky TQ and Ben-Hamu, Heli and Nickel, Maximilian and Le, Matt},
  journal={arXiv preprint arXiv:2210.02747},
  year={2022}
}

@inproceedings{wang2025language,
  title={Language model based text-to-audio generation: Anti-causally aligned collaborative residual transformers},
  author={Wang, Juncheng and Xu, Chao and Yu, Cheng and Hu, Zhe and Xie, Haoyu and Yu, Guoqi and Shang, Lei and Wang, Shujun},
  booktitle={Proceedings of the 2025 Conference on Empirical Methods in Natural Language Processing},
  pages={26036--26054},
  year={2025}
}

@article{linapoavatar,
  title={ApoAvatar: Expressive Audio-Driven Avatar Generation via Refocused Audio-Pose Priors},
  author={Lin, Jingyu and Zhang, Chao and Feng, Wei and Zhou, Donghao and Wen, Shilei and Du, Lan and Chen, Cunjian}
}

@article{zhou2024magictailor,
  title={Magictailor: Component-controllable personalization in text-to-image diffusion models},
  author={Zhou, Donghao and Huang, Jiancheng and Bai, Jinbin and Wang, Jiaze and Chen, Hao and Chen, Guangyong and Hu, Xiaowei and Heng, Pheng-Ann},
  journal={arXiv preprint arXiv:2410.13370},
  year={2024}
}

@article{song2025scenedecorator,
  title={SceneDecorator: Towards Scene-Oriented Story Generation with Scene Planning and Scene Consistency},
  author={Song, Quanjian and Zhou, Donghao and Lin, Jingyu and Shen, Fei and Wang, Jiaze and Hu, Xiaowei and Chen, Cunjian and Heng, Pheng-Ann},
  journal={arXiv preprint arXiv:2510.22994},
  year={2025}
}

@article{song2025hero,
  title={HERO: Hierarchical Extrapolation and Refresh for Efficient World Models},
  author={Song, Quanjian and Wang, Xinyu and Zhou, Donghao and Lin, Jingyu and Chen, Cunjian and Ma, Yue and Li, Xiu},
  journal={arXiv preprint arXiv:2508.17588},
  year={2025}
}

@inproceedings{huang2025dual,
  title={Dual-Schedule Inversion: Training-and Tuning-Free Inversion for Real Image Editing},
  author={Huang, Jiancheng and Huang, Yi and Liu, Jianzhuang and Zhou, Donghao and Liu, Yifan and Chen, Shifeng},
  booktitle={2025 IEEE/CVF Winter Conference on Applications of Computer Vision (WACV)},
  pages={660--669},
  year={2025},
  organization={IEEE}
}

@inproceedings{zhou2026identitystory,
  title={IdentityStory: Taming Your Identity-Preserving Generator for Human-Centric Story Generation},
  author={Zhou, Donghao and Lin, Jingyu and Shen, Guibao and Liu, Quande and Gao, Jialin and Liu, Lihao and Du, Lan and Chen, Cunjian and Fu, Chi-Wing and Hu, Xiaowei and others},
  booktitle={Proceedings of the AAAI Conference on Artificial Intelligence},
  volume={40},
  number={16},
  pages={13593--13601},
  year={2026}
}

@inproceedings{tan2025ominicontrol,
  title={Ominicontrol: Minimal and universal control for diffusion transformer},
  author={Tan, Zhenxiong and Liu, Songhua and Yang, Xingyi and Xue, Qiaochu and Wang, Xinchao},
  booktitle={Proceedings of the IEEE/CVF International Conference on Computer Vision},
  pages={14940--14950},
  year={2025}
}

@inproceedings{ju2025fulldit,
  title={FullDiT: Video Generative Foundation Models with Multimodal Control via Full Attention},
  author={Ju, Xuan and Ye, Weicai and Liu, Quande and Wang, Qiulin and Wang, Xintao and Wan, Pengfei and Zhang, Di and Gai, Kun and Xu, Qiang},
  booktitle={Proceedings of the IEEE/CVF International Conference on Computer Vision},
  pages={15737--15747},
  year={2025}
}

@article{baevski2020wav2vec,
  title={wav2vec 2.0: A framework for self-supervised learning of speech representations},
  author={Baevski, Alexei and Zhou, Yuhao and Mohamed, Abdelrahman and Auli, Michael},
  journal={Advances in neural information processing systems},
  volume={33},
  pages={12449--12460},
  year={2020}
}

@inproceedings{zhang2023adding,
  title={Adding conditional control to text-to-image diffusion models},
  author={Zhang, Lvmin and Rao, Anyi and Agrawala, Maneesh},
  booktitle={Proceedings of the IEEE/CVF international conference on computer vision},
  pages={3836--3847},
  year={2023}
}

@inproceedings{weimocha,
  title={MoCha: Towards Movie-Grade Talking Character Generation},
  author={Wei, Cong and Sun, Bo and Ma, Haoyu and Hou, Ji and Juefei-Xu, Felix and He, Zecheng and Dai, Xiaoliang and Zhang, Luxin and Li, Kunpeng and Hou, Tingbo and others},
  booktitle={The Thirty-ninth Annual Conference on Neural Information Processing Systems}
}

@article{liu2026hifi,
  title={HiFi-Inpaint: Towards High-Fidelity Reference-Based Inpainting for Generating Detail-Preserving Human-Product Images},
  author={Liu, Yichen and Zhou, Donghao and Wang, Jie and Gao, Xin and Liu, Guisheng and Li, Jiatong and Zhang, Quanwei and Lyu, Qiang and Guo, Lanqing and Wen, Shilei and others},
  journal={arXiv preprint arXiv:2603.02210},
  year={2026}
}

@inproceedings{huang2024magicfight,
  title={Magicfight: Personalized martial arts combat video generation},
  author={Huang, Jiancheng and Yan, Mingfu and Chen, Songyan and Huang, Yi and Chen, Shifeng},
  booktitle={Proceedings of the 32nd ACM International Conference on Multimedia},
  pages={10833--10842},
  year={2024}
}

@article{huang2025m4v,
  title={M4V: Multi-Modal Mamba for Text-to-Video Generation},
  author={Huang, Jiancheng and Zhang, Gengwei and Jie, Zequn and Jiao, Siyu and Qian, Yinlong and Chen, Ling and Wei, Yunchao and Ma, Lin},
  journal={arXiv preprint arXiv:2506.10915},
  year={2025}
}

@article{loshchilov2017decoupled,
  title={Decoupled weight decay regularization},
  author={Loshchilov, Ilya and Hutter, Frank},
  journal={arXiv preprint arXiv:1711.05101},
  year={2017}
}

@inproceedings{chung2016out,
  title={Out of time: automated lip sync in the wild},
  author={Chung, Joon Son and Zisserman, Andrew},
  booktitle={Asian conference on computer vision},
  pages={251--263},
  year={2016},
  organization={Springer}
}

@article{yuan2025opens2v,
  title={Opens2v-nexus: A detailed benchmark and million-scale dataset for subject-to-video generation},
  author={Yuan, Shenghai and He, Xianyi and Deng, Yufan and Ye, Yang and Huang, Jinfa and Lin, Bin and Luo, Jiebo and Yuan, Li},
  journal={arXiv preprint arXiv:2505.20292},
  year={2025}
}

@inproceedings{huang2024vbench,
  title={Vbench: Comprehensive benchmark suite for video generative models},
  author={Huang, Ziqi and He, Yinan and Yu, Jiashuo and Zhang, Fan and Si, Chenyang and Jiang, Yuming and Zhang, Yuanhan and Wu, Tianxing and Jin, Qingyang and Chanpaisit, Nattapol and others},
  booktitle={Proceedings of the IEEE/CVF Conference on Computer Vision and Pattern Recognition},
  pages={21807--21818},
  year={2024}
}

@inproceedings{yang2023effective,
  title={Effective whole-body pose estimation with two-stages distillation},
  author={Yang, Zhendong and Zeng, Ailing and Yuan, Chun and Li, Yu},
  booktitle={Proceedings of the IEEE/CVF International Conference on Computer Vision},
  pages={4210--4220},
  year={2023}
}

@article{liu2025improving,
  title={Improving video generation with human feedback},
  author={Liu, Jie and Liu, Gongye and Liang, Jiajun and Yuan, Ziyang and Liu, Xiaokun and Zheng, Mingwu and Wu, Xiele and Wang, Qiulin and Xia, Menghan and Wang, Xintao and others},
  journal={arXiv preprint arXiv:2501.13918},
  year={2025}
}

@misc{google_nanobanana,
  author       = {Google},
  title        = {Nano Banana},
  year         = {2026},
  month        = {January},
  url          = {https://ai.google.dev/gemini-api/docs/image-generation},
  note         = {Accessed: 2026-01-26},
  publisher    = {Google AI for Developers},
  howpublished = {Online Documentation}
}

@article{hurst2024gpt,
  title={Gpt-4o system card},
  author={Hurst, Aaron and Lerer, Adam and Goucher, Adam P and Perelman, Adam and Ramesh, Aditya and Clark, Aidan and Ostrow, AJ and Welihinda, Akila and Hayes, Alan and Radford, Alec and others},
  journal={arXiv preprint arXiv:2410.21276},
  year={2024}
}

@misc{elevenlabs_ai_voice,
  author       = {ElevenLabs},
  title        = {ElevenLabs: The most realistic voice AI platform},
  year         = {2026},
  url          = {https://elevenlabs.io/},
  note         = {Accessed: 2026-01-26},
  publisher    = {ElevenLabs},
  howpublished = {Online Platform}
}

@misc{pyscenedetect,
  author       = {Brandon Castellano},
  title        = {PySceneDetect: Python and OpenCV-based Scene Cut/Transition Detection Program \& Library},
  year         = {2025},
  url          = {https://github.com/Breakthrough/PySceneDetect},
  note         = {Accessed: 2026-01-26},
  publisher    = {GitHub},
  howpublished = {Online Repository}
}

@inproceedings{esser2024scaling,
  title={Scaling rectified flow transformers for high-resolution image synthesis},
  author={Esser, Patrick and Kulal, Sumith and Blattmann, Andreas and Entezari, Rahim and M{\"u}ller, Jonas and Saini, Harry and Levi, Yam and Lorenz, Dominik and Sauer, Axel and Boesel, Frederic and others},
  booktitle={Forty-first international conference on machine learning},
  year={2024}
}

@article{su2024roformer,
  title={Roformer: Enhanced transformer with rotary position embedding},
  author={Su, Jianlin and Ahmed, Murtadha and Lu, Yu and Pan, Shengfeng and Bo, Wen and Liu, Yunfeng},
  journal={Neurocomputing},
  volume={568},
  pages={127063},
  year={2024},
  publisher={Elsevier}
}

\newpage
\appendix
\onecolumn

\section{Training Data Collection}

\label{sec:data_details}

\begin{figure*}[h]
    \centering
    
    \vspace{2mm}

    \includegraphics[width=1\linewidth]{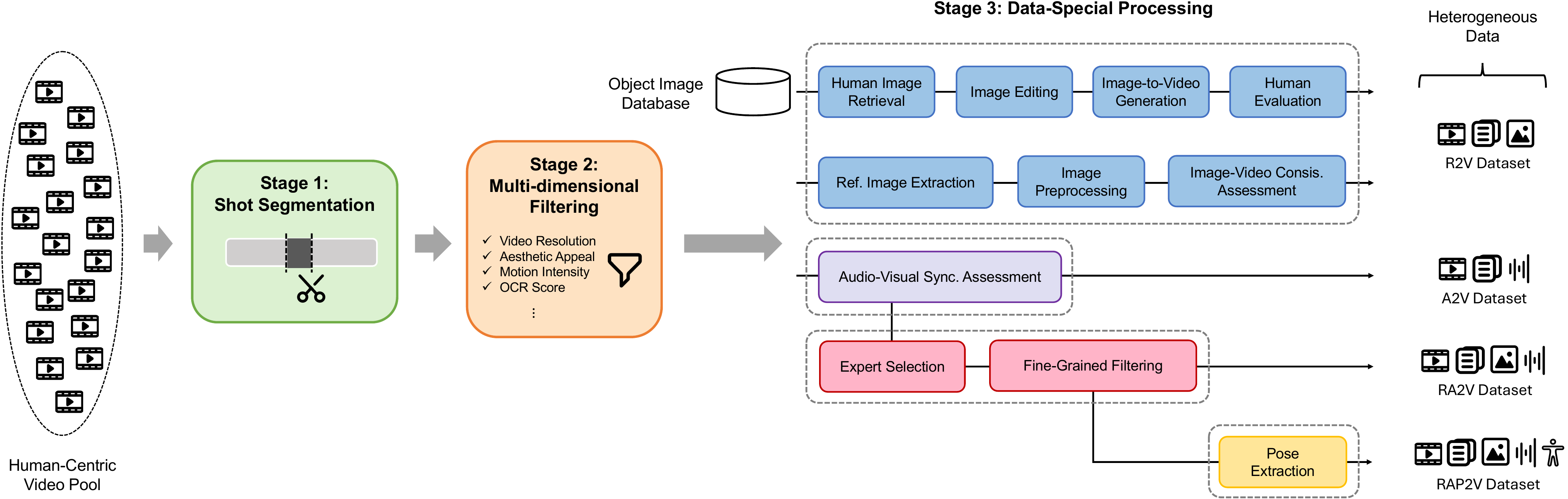}

    \vspace{2mm}

    \caption
    { 
        \textbf{Data collection pipeline.}
        We first segment videos into isolated shots and then filter them by diverse metric scores. Subsequently, specialized processing workflows are applied to construct multiple heterogeneous datasets.
    }

    \vspace{2mm}

    \label{fig:data_prep}
    
\end{figure*}

A video generation model's potential is mainly bounded by the richness, diversity, and scale of the data upon which it is trained. 
However, high-quality data for Human-Object Interaction Video Generation (HOIVG) is scarce, as it necessitates the simultaneous validity of text, reference images, audio, and pose conditions. To address this challenge, we construct a large-scale, heterogeneous dataset through a rigorous pipeline designed to meet specific sub-task standards. As illustrated in \cref{fig:data_prep}, our pipeline consists of three main stages.

\textbf{Stage 1: Shot Segmentation.}
We begin by collecting a massive in-house human-centric video pool. To ensure temporal coherence and content focus, we apply a shot segmentation algorithm \cite{pyscenedetect} to decompose long videos into individual clips. This process isolates continuous shots, effectively removing scene transitions and ensuring that each clip contains a single, uninterrupted visual narrative.

\textbf{Stage 2: Multi-dimensional Filtering.}
Following segmentation, we implement a filtering process to rigorously remove low-quality samples. We employ a multi-dimensional criteria set to screen the clips based on various metrics, such as (1) \textit{video resolution}, ensuring high visual fidelity; (2) \textit{aesthetic appeal}, filtering for visually pleasing composition and lighting; (3) \textit{motion intensity}, selecting clips with sufficient movement to facilitate effective motion learning; and (4) \textit{OCR score}, removing clips with excessive on-screen text or watermarks that might degrade generation quality. This comprehensive filtering significantly reduces noise and ensures that only high-quality candidates proceed to the next phase.

\textbf{Stage 3: Data-Special Processing.}
To maximize data utilization and support various training objectives, we process the filtered clips into four distinct heterogeneous datasets, each tailored for specific multimodal condition training (for notational conciseness, we omit ``text'' from respective task names):

\begin{itemize}[itemsep=-0.2em, topsep=-0.2em]
\item \textbf{R2V (Reference-to-Video) Dataset:} For synthesized data, we utilize an internal object image database. The process involves retrieving relevant human images, performing image editing to compose human and object concepts, and generating corresponding videos via an internal image-to-video generation model. To ensure quality, we incorporate a human evaluation step, where crowd experts filter out samples exhibiting distortions or excessive ``AI-ness'' (\eg, loss of fine details or over-smoothing). 
Additionally, we extract reference images from real videos, perform preprocessing such as image super-resolution, and conduct image-video consistency assessment to guarantee alignment.
\item \textbf{A2V (Audio-to-Video) Dataset:} This dataset focuses on audio-driven generation. We perform audio-visual synchronization assessment to select clips where the audio (mainly speech) aligns perfectly with the visual actions, discarding samples with significant misalignment.

\item \textbf{RA2V (Reference+Audio-to-Video) Dataset:} As a high-quality dataset for joint training, this dataset undergoes the most stringent curation. It combines the criteria of the previous datasets and adds an expert selection phase followed by fine-grained filtering. Each video clip is independently reviewed to ensure it meets the highest standards of visual quality and semantic consistency across both reference image and audio modalities.

\item \textbf{RAP2V (Reference+Audio+Pose-to-Video) Dataset:} Finally, for the final fine-tuning, we extend a high-quality subset of the RA2V dataset to apply pose extraction. We extract per-frame human pose skeletons from the validated videos, providing ground truth necessary for precise motion control.
\end{itemize}

By further integrating video captioning to provide comprehensive descriptions of human subjects, objects, actions, environments, and interaction details, this pipeline successfully curates a total of $\bm{\mathcal{O}(1\text{\textbf{m}})}$ \textbf{clips}, amounting to around \textbf{3500 hours}. This collection serves as a robust foundation for training our HOIVG model.

\section{More Implementation Details}
\label{sec:more_impl_details}

Our model is built upon the 12B Waver 1.0 model \cite{zhang2025waver}, which utilizes the Multimodal Diffusion Transformer (MMDiT) architecture \cite{esser2024scaling}. The training framework employs Flow Matching \cite{lipman2022flow} to supervise video sequences. The entire training process involves an initial phase using 480p videos followed by a high-resolution phase using 720p videos. To manage the computational demands of long video training samples (up to 241 frames at 24 fps), we set the Ulysses-style sequence parallelism \cite{jacobs2023deepspeed} size to 8 within our distributed setup. Furthermore, to maximize training efficiency, all features are extracted offline before training. 
Notably, incorporating the proposed \techAudio increases the model scale by merely 0.3B parameters. This efficiency is attributed to the insights provided by the adaptive gating mechanism, as shown in \cref{fig:gating_insight}. In summary, our approach achieves high-quality video generation while maintaining a highly efficient parameter footprint relative to the 12B backbone.

\begin{table*}[t]
    \caption{
    \textbf{Quantitative comparison with the cascaded baseline on the RAP2V task.}
    We compare our end-to-end \method with a representative cascaded baseline (VACE~\cite{jiang2025vace}+LatentSync~\cite{li2024latentsync}). Our approach outperforms the cascaded baseline across all evaluation metrics, highlighting the superiority of multi-condition unification.
    }
    
    \centering
    \setlength{\tabcolsep}{2.5mm}
    \renewcommand{\arraystretch}{1.2}
    
    \resizebox{\linewidth}{!}
    {
        \begin{tabular}{lccccccccccc}
        \specialrule{0.1em}{0pt}{2pt}
        \multirow{2}[4]{*}{Method} & Text Align. & \multicolumn{2}{c}{Reference Consistency} & \multicolumn{2}{c}{Audio-Visual Sync.} & \multicolumn{2}{c}{Pose Accuracy} & \multicolumn{4}{c}{Video Quality} \\
        \cmidrule(lr){2-2} \cmidrule(lr){3-4} \cmidrule(lr){5-6} \cmidrule(lr){7-8} \cmidrule(lr){9-12}
              & TA$\uparrow$ & FaceSim$\uparrow$ & NexusScore$\uparrow$ & Sync-C$\uparrow$ & Sync-D$\downarrow$ & AKD$\downarrow$ & PCK$\uparrow$ & AES$\uparrow$ & IQA$\uparrow$ & VQ$\uparrow$ & MQ$\uparrow$ \\
        \midrule
        Cascaded Baseline & 6.885 & 0.591 & 0.341 & 7.016 & 7.823 & 0.198 & 0.340 & 0.417 & 0.709 & 10.05 & 3.911 \\
        \method (Ours) & \textbf{7.134} & \textbf{0.645} & \textbf{0.353} & \textbf{7.699} & \textbf{7.674} & \textbf{0.172} & \textbf{0.478} & \textbf{0.424} & \textbf{0.725} & \textbf{11.06} & \textbf{5.880} \\
        \specialrule{0.1em}{1pt}{0pt}
        \end{tabular}
    } 
    \label{tab:quant_cascaded}
\end{table*}

\begin{figure*}[t]
    \centering
    
    \includegraphics[width=1\linewidth]{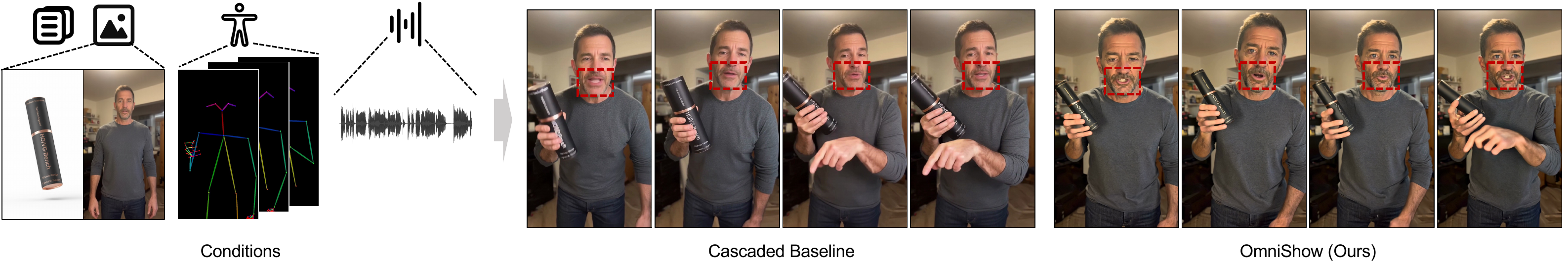}
    
    \caption{
        \textbf{Qualitative comparison with the cascaded baseline on the RAP2V task.} 
        Compared to the cascaded baseline (VACE~\cite{jiang2025vace}+LatentSync~\cite{li2024latentsync}), our \method generates visually coherent videos with precise lip synchronization and superior visual quality. Notably, it avoids the artifacts and blur often introduced by isolated lip-synchronization generation, particularly under complex occlusions. \textit{Zoom in for better view.}
    }
    \label{fig:qual_cascaded}
\end{figure*}

\section{Additional Comparison}
\label{sec:more_comp}

To further validate the effectiveness of our end-to-end multi-condition unification, we construct a cascaded baseline for the RAP2V task. Specifically, this baseline sequentially combines VACE~\cite{jiang2025vace} (for RP2V generation) with LatentSync~\cite{li2024latentsync} (for audio-driven lip-synchronization). 
We believe it serves as a representative approach to sequentially synthesize RAP2V samples.

As shown in the quantitative results in \cref{tab:quant_cascaded}, the cascaded baseline struggles across multiple metrics. In contrast, our \method achieves superior performance across the board. Notably, our end-to-end approach yields better audio-visual synchronization (\eg, a Sync-C of 7.699 compared to 7.016) and pose accuracy (\eg, a PCK of 0.478 compared to 0.340). Furthermore, video and motion quality metrics are also substantially higher (such as a VQ of 11.06 versus 10.05), highlighting the effectiveness of simultaneously modeling all conditions.
Furthermore, we provide qualitative comparisons in \cref{fig:qual_cascaded}. \method consistently generates highly coherent videos with better visual fidelity compared to the cascaded baseline. Specifically, conducting isolated lip-synchronization generation as a post-processing step in the cascaded baseline leads to visual artifacts and blur, especially under mouth occlusions, which can be attributed to the inherent limitations of non-end-to-end approaches. This comparison strongly highlights the necessity and advantages of our unified end-to-end framework for HOIVG.

\section{Additional Ablation Studies}
\label{sec:more_abl}

\begin{table*}[t]
    \caption{\textbf{Quantitative results of additional ablation studies.} We further evaluate two design choices within the proposed techniques, including 
    (a) different RoPE strategies for positional encoding of the pseudo-frames in \techTrain and 
    (b) The impact of context window size in \techAudio.}
    \label{tab:additional_ablation}
    
    \centering
    \renewcommand{\arraystretch}{1.1}

    \begin{subtable}{0.46\linewidth}
        \begin{minipage}[t]{0.08\linewidth}
            \vspace{0pt} 
            {\small (a)} 
        \end{minipage}
        \begin{minipage}[t]{0.90\linewidth}
            \vspace{0pt} 
            \setlength{\tabcolsep}{1.7mm} 
            \resizebox{\linewidth}{!}{
                \begin{tabular}[t]{lccc} 
                \toprule
                RoPE Strategy & FaceSim$\uparrow$ & NexusScore$\uparrow$ & AES$\uparrow$ \\
                \midrule
                Temporal Shift & \underline{0.675} & \underline{0.351} & \underline{0.468} \\
                Spatiotemporal Shift & 0.279 & 0.339 & 0.456 \\
                Native (Ours) & \textbf{0.707} & \textbf{0.353} & \textbf{0.471} \\
                \bottomrule
                \end{tabular}
            }
        \end{minipage}
        \phantomcaption\label{tab:abl_rope}
    \end{subtable}
    \hfill
    \begin{subtable}{0.48\linewidth}
        \begin{minipage}[t]{0.08\linewidth}
            \vspace{0pt}
            {\small (b)}
        \end{minipage}
        \begin{minipage}[t]{0.90\linewidth}
            \vspace{0pt}
            \setlength{\tabcolsep}{1.7mm}
            \resizebox{\linewidth}{!}{
                \begin{tabular}[t]{lccc}
                \toprule
                Context Setup & Sync-C$\uparrow$ & Sync-D$\downarrow$ & AES$\uparrow$ \\
                \midrule
                Context Window = 1 & \underline{8.872} & \underline{7.878} & \underline{0.533} \\
                Context Window = 11 & 7.020 & 9.588 & 0.527 \\
                Context Window = 5 (Ours) & \textbf{9.023} & \textbf{7.419} & \textbf{0.540} \\
                \bottomrule
                \end{tabular}
            }
        \end{minipage}
        \phantomcaption\label{tab:abl_context}
    \end{subtable}
    
\end{table*}

\subsection{Position Embeddings for \techVision}
Modern MMDiT models use 3D Spatiotemporal Rotary Positional Embeddings (RoPE) \cite{su2024roformer}.  As for the RoPE setup of pseudo-frames, we adopt the native strategy, treating pseudo-frames and original video frames as a continuous sequence starting from $T=0$. We compared this against a ``Temporal Shift'' strategy (using negative indices like $T=-1,-2$) and a ``Spatiotemporal Shift'' strategy (adding spatial offsets) following \cite{hu2025hunyuancustom}. As shown in \cref{tab:abl_rope}, empirical results in the R2V task favor our native strategy. We attribute this to the model's pre-training on standard continuous video. Artificial shifts create mismatches for the channel-wise conditioning, whereas our approach aligns with the model's inherent expectation of temporal continuity, allowing it to effectively utilize the reference context.

\subsection{Window Size of Audio Context Packing}
The proposed \techAudio employs a sliding window of $w=5$ to capture phonetic context. We validated this against $w=1$ (\ie, no audio context) and $w=11$. As illustrated in \cref{tab:abl_context}, quantitative results show $w=5$ offers the best synchronization. A small window ($w=1$) fails to capture temporal correlation, resulting in jittery transitions between phonemes and overreaction. Conversely, an excessive window ($w=11$) causes ``over-smoothing,'' where broad context dilutes the fine-grained cues needed for precise lip synchronization. Thus, $w=5$ strikes the optimal balance, providing sufficient context without obscuring the instantaneous audio signal required for precise synchronization.

\section{More Qualitative Results}

\label{sec:more_qual_results}

\begin{figure*}[t!]
    \centering
    
    \includegraphics[width=1\linewidth]{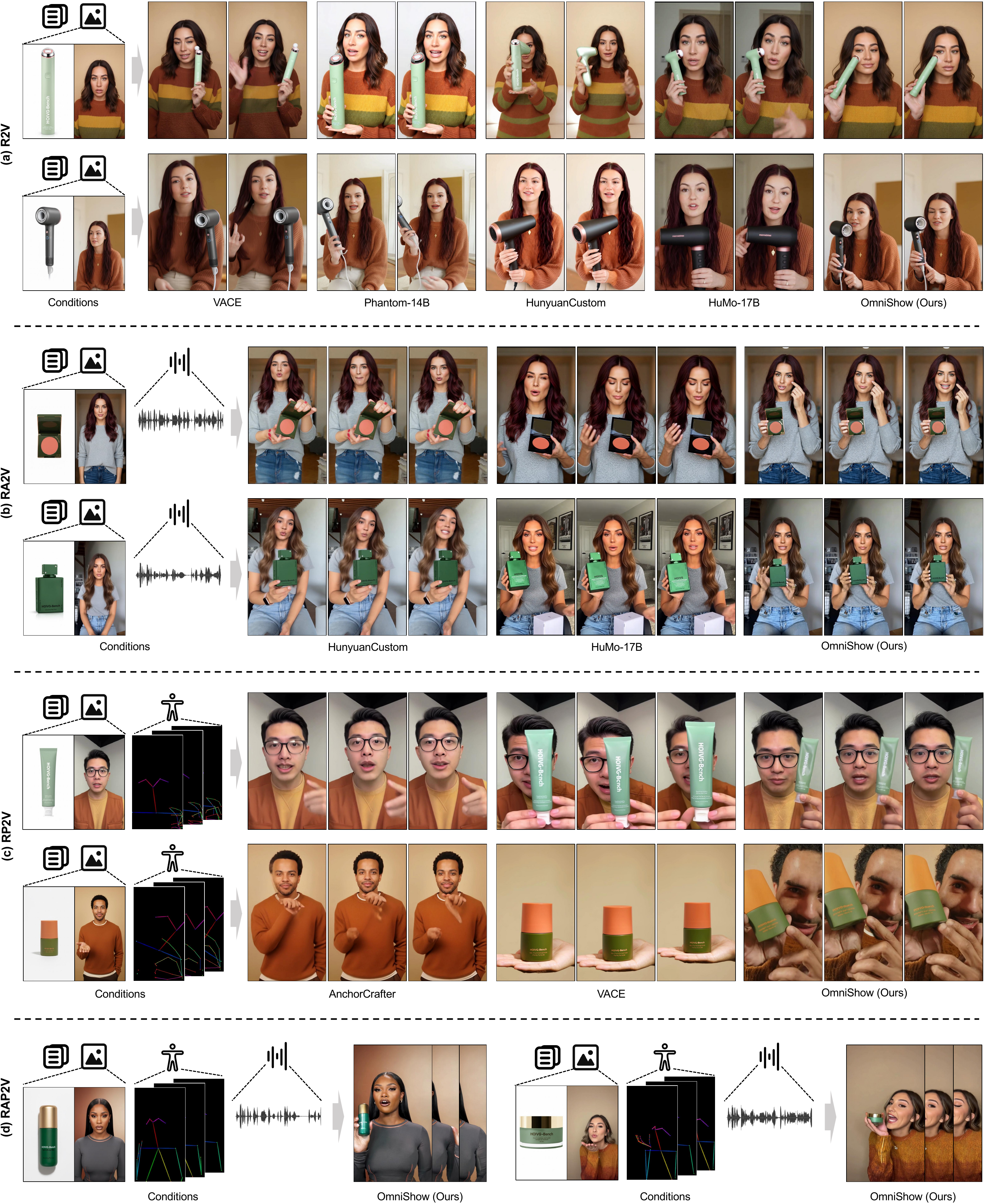}
    
    \caption
    { 
        \textbf{More qualitative results.}
       We present additional generated results from our \method and other methods across various multimodal condition settings, showing the state-of-the-art performance of \method. Note that \method is the first-of-its-kind framework supporting the full spectrum of four multimodal inputs required by HOIVG.
       \textit{We recommend visiting our project page for video demonstrations.}
    }
    
    \label{fig:qual_results_more}
    
\end{figure*}

\begin{figure*}[t!]
    \centering
    
    \includegraphics[width=1\linewidth]{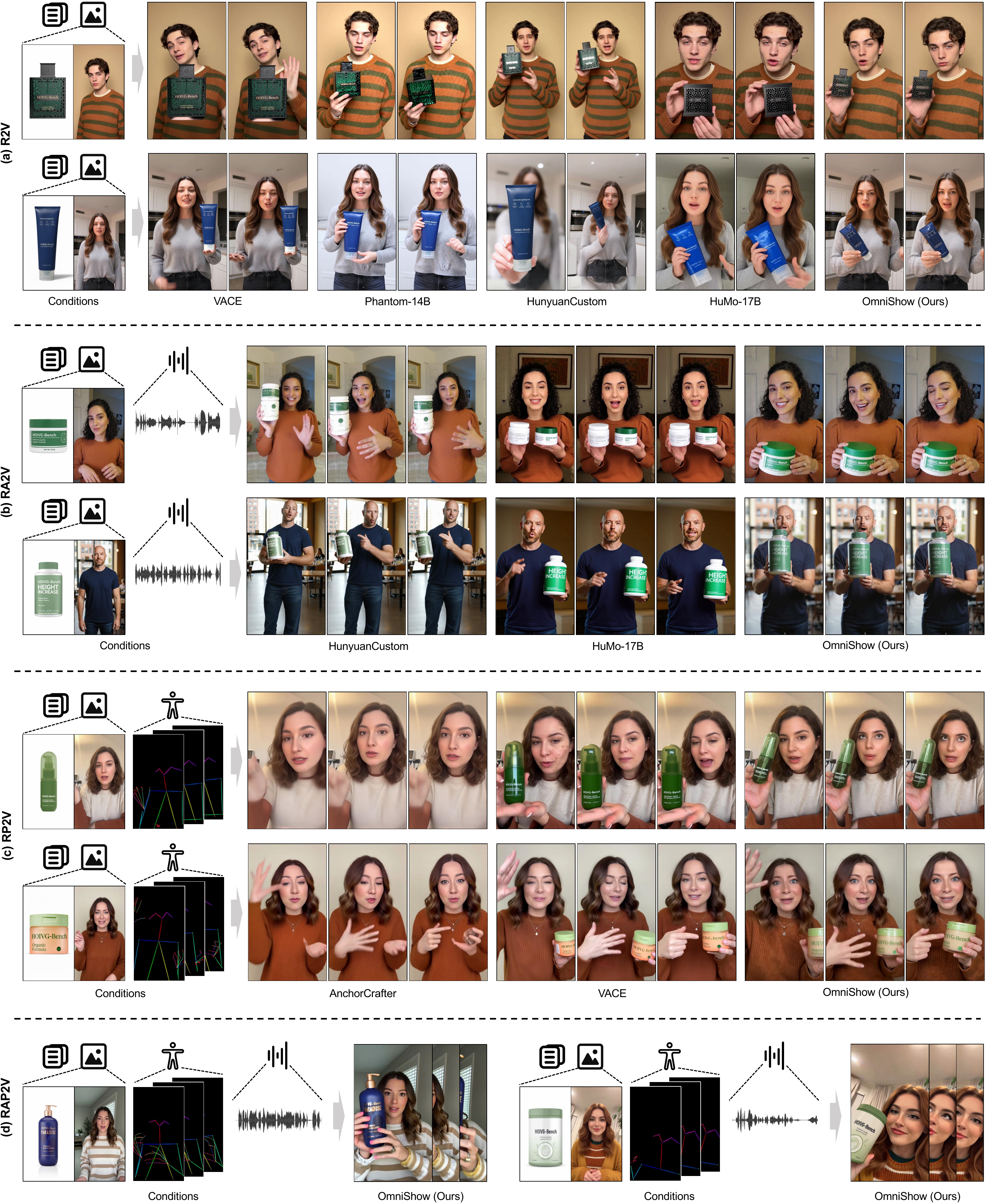}
    
    \caption
    { 
        \textbf{More qualitative results.}
       We present additional generated results from our \method and other methods across various multimodal condition settings, showing the state-of-the-art performance of \method. Note that \method is the first-of-its-kind framework supporting the full spectrum of four multimodal inputs required by HOIVG.
       \textit{We recommend visiting our project page for video demonstrations.}
    }
    
    \label{fig:qual_results_more_2}
    
\end{figure*}

\begin{figure*}[t!]
    \centering
    
    \includegraphics[width=1\linewidth]{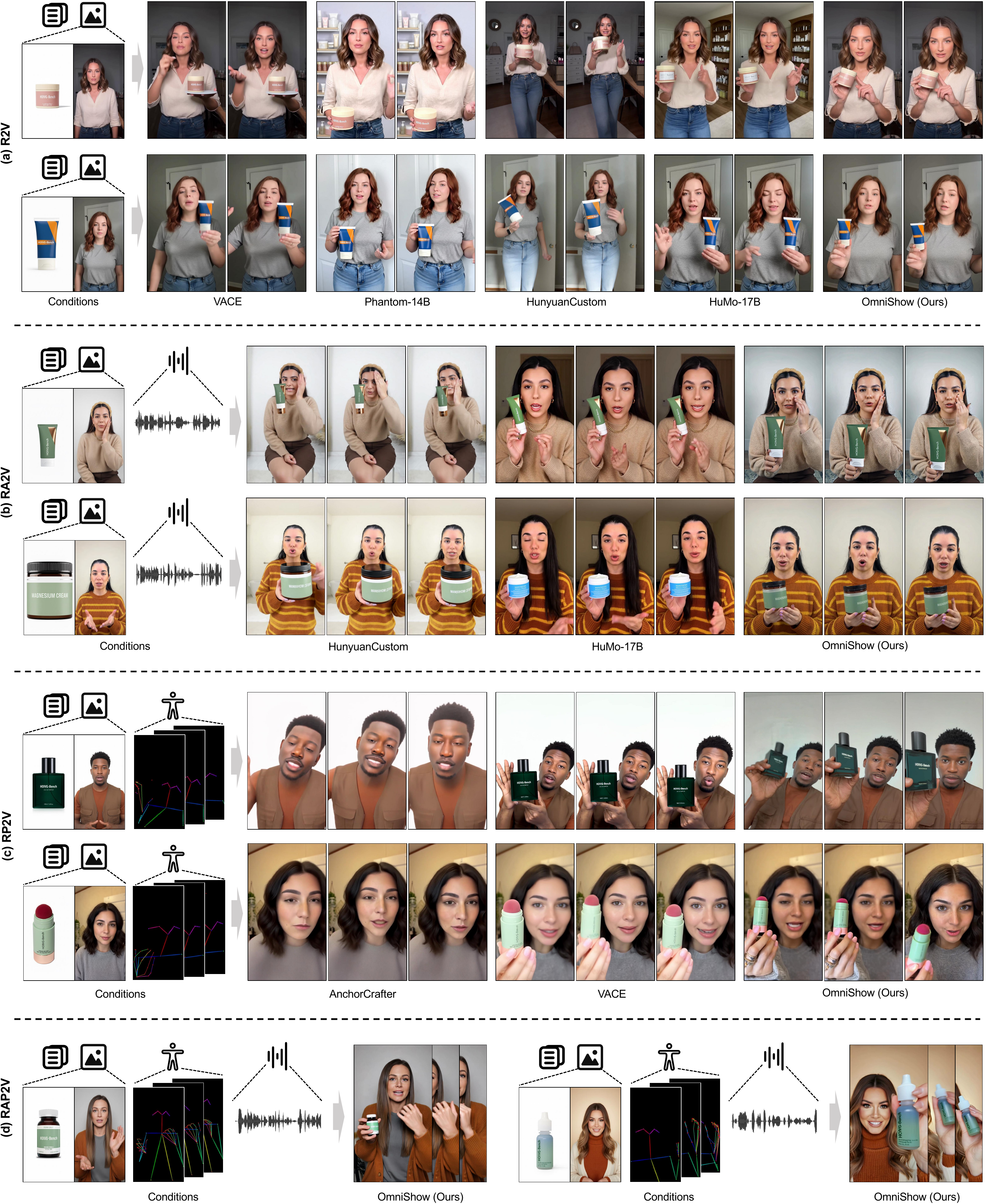}
    
    \caption
    { 
        \textbf{More qualitative results.}
       We present additional generated results from our \method and other methods across various multimodal condition settings, showing the state-of-the-art performance of \method. Note that \method is the first-of-its-kind framework supporting the full spectrum of four multimodal inputs required by HOIVG.
       \textit{We recommend visiting our project page for video demonstrations.}
    }
    
    \label{fig:qual_results_more_3}
    
\end{figure*}

To provide a substantially more comprehensive and detailed assessment of the generative capabilities of \method, we present an extensive set of additional qualitative results in \cref{fig:qual_results_more}, \cref{fig:qual_results_more_2}, and \cref{fig:qual_results_more_3}. 
These visualization results cover a highly diverse and challenging range of multimodal conditioning settings, including R2V, RA2V, RP2V, and RAP2V. 

Across all of these varied and demanding scenarios, our proposed model consistently exhibits superior generation quality, which is characterized by high-fidelity reference identity preservation, natural motion dynamics, and precise audio-visual synchronization. 
Most notably, the provided additional samples further serve to vividly illustrate the model's exceptional robustness in effectively handling complex spatial human-object interactions as well as large pose variations, without compromising underlying visual details. 
Furthermore, we showcase the unique capability of our unified framework by presenting compelling results under the fully combined conditioning setting (\ie, RAP2V), conclusively verifying its profound effectiveness in seamlessly orchestrating all input modalities simultaneously to generate highly coherent, temporally smooth, and richly expressive videos.
Video demonstrations are available on our project page.


\end{document}